\documentclass{article}

\usepackage{microtype}
\usepackage{graphicx}
\usepackage{subcaption}
\usepackage{placeins}
\usepackage{booktabs} 

\usepackage{hyperref}
\usepackage{tablefootnote}

\usepackage[accepted]{icml2026}

\usepackage{amsmath}
\usepackage{amssymb}
\usepackage{mathtools}
\usepackage{amsthm}

\usepackage{tikz}
\usetikzlibrary{shapes, arrows.meta, positioning, shadows, fit, calc}

\usepackage[capitalize,noabbrev]{cleveref}
\usepackage{listings}
\usepackage{multirow}
\usepackage[most]{tcolorbox}
\tcbuselibrary{listings}
\usepackage{xcolor}
\usepackage{amsthm}
\newtheorem*{remark}{Remark}

\definecolor{darkgray}{RGB}{60, 60, 60}
\definecolor{sophisticatedbg}{RGB}{242, 247, 255}
\definecolor{burntorange}{RGB}{204, 85, 0}

\newtcblisting{promptbox}[2][]{%
    colback=gray!5,
    colframe=darkgray,
    colbacktitle=darkgray,
    coltitle=white,
    fonttitle=\bfseries\sffamily,
    title={#2},
    arc=2mm,
    boxrule=1pt,
    left=2mm, right=2mm, top=2mm, bottom=2mm,
    breakable,
    enhanced,
    listing only,        
    listing options={    
        basicstyle=\footnotesize\ttfamily,
        breaklines=true,       
        breakatwhitespace=false,
        columns=fullflexible,  
        keepspaces=true,
        tabsize=2,
        showstringspaces=false 
    },
    #1
}

\newtcblisting{resultbox}[2][]{%
    colback=sophisticatedbg,       
    colframe=blue!30!black,        
    colbacktitle=blue!30!black,    
    coltitle=white,                
    fonttitle=\bfseries\sffamily,
    title={#2},
    arc=2mm, boxrule=0.5pt,
    left=2mm, right=2mm, top=2mm, bottom=2mm,
    breakable, enhanced,
    listing only,
    #1
}

\newtcolorbox{questionbox}[2][]{%
    colback=burntorange!10,    
    colframe=burntorange,      
    colbacktitle=burntorange,  
    coltitle=white,            
    fonttitle=\bfseries\sffamily,
    title={#2},
    arc=2mm,
    boxrule=1pt,
    left=3mm, right=3mm, top=3mm, bottom=3mm,
    breakable,
    enhanced,
    #1
}

\theoremstyle{plain}

\theoremstyle{definition}

\theoremstyle{remark}

\icmltitlerunning{NEMO: Execution-Aware Optimization Modeling via Autonomous Coding Agents}

\begin{document}

\twocolumn[
  \icmltitle{NEMO: Execution-Aware Optimization Modeling via Autonomous Coding Agents}



  \icmlsetsymbol{equal}{*}

  \begin{icmlauthorlist}
    \icmlauthor{Yang Song}{equal,c3ai}
    \icmlauthor{Anoushka Vyas}{equal,c3ai}
    \icmlauthor{Zirui Wei}{c3ai}
    \icmlauthor{Sina Khoshfetrat Pakazad}{c3ai}
    \icmlauthor{Henrik Ohlsson}{c3ai}
    \icmlauthor{Graham Neubig}{cmu}
  \end{icmlauthorlist}

  \icmlaffiliation{c3ai}{C3 AI, Redwood City, CA, USA}
  \icmlaffiliation{cmu}{Language Technologies Institute, School of Computer Science, Carnegie Mellon University, Pittsburgh, PA, USA}

  \icmlcorrespondingauthor{Sina Khoshfetrat Pakazad}{sina.pakazad@c3.ai}

    \icmlkeywords{Autonomous Coding Agents, Optimization Modeling, Operations Research, Large Language Models, LLM Agents, Code Generation, Execution-Aware Validation}

  \vskip 0.3in
]



\printAffiliationsAndNotice{\icmlEqualContribution}  

\begin{abstract}

We present \textbf{NEMO}, a system that translates \textbf{N}atural-language descriptions of decision problems into formal \textbf{E}xecutable \textbf{M}athematical \textbf{O}ptimization implementations using autonomous coding agents (ACAs). Existing approaches rely on specialized large language models (LLMs) or bespoke task-specific agents that are often brittle and frequently generate syntactically invalid or non-executable code. NEMO instead treats ACAs as a first-class abstraction analogous to API-based interaction with LLMs; their sandboxed execution guarantees code is executable by construction and supports automated validation and repair. We introduce novel coordination patterns including asymmetric validation loops between independently generated optimizer and simulator implementations, external memory for experience reuse, and robustness enhancements via minimum Bayes risk (MBR) decoding and self-consistency. Across nine established optimization benchmarks, NEMO achieves state-of-the-art performance on the majority of tasks with substantial margins on several datasets (Figure~\ref{fig:sota_comparison}), demonstrating the power of execution-aware agentic architectures for automated optimization modeling.
\end{abstract}

\begin{figure}[t]
  \begin{center}
    \centerline{\includegraphics[width=\linewidth]{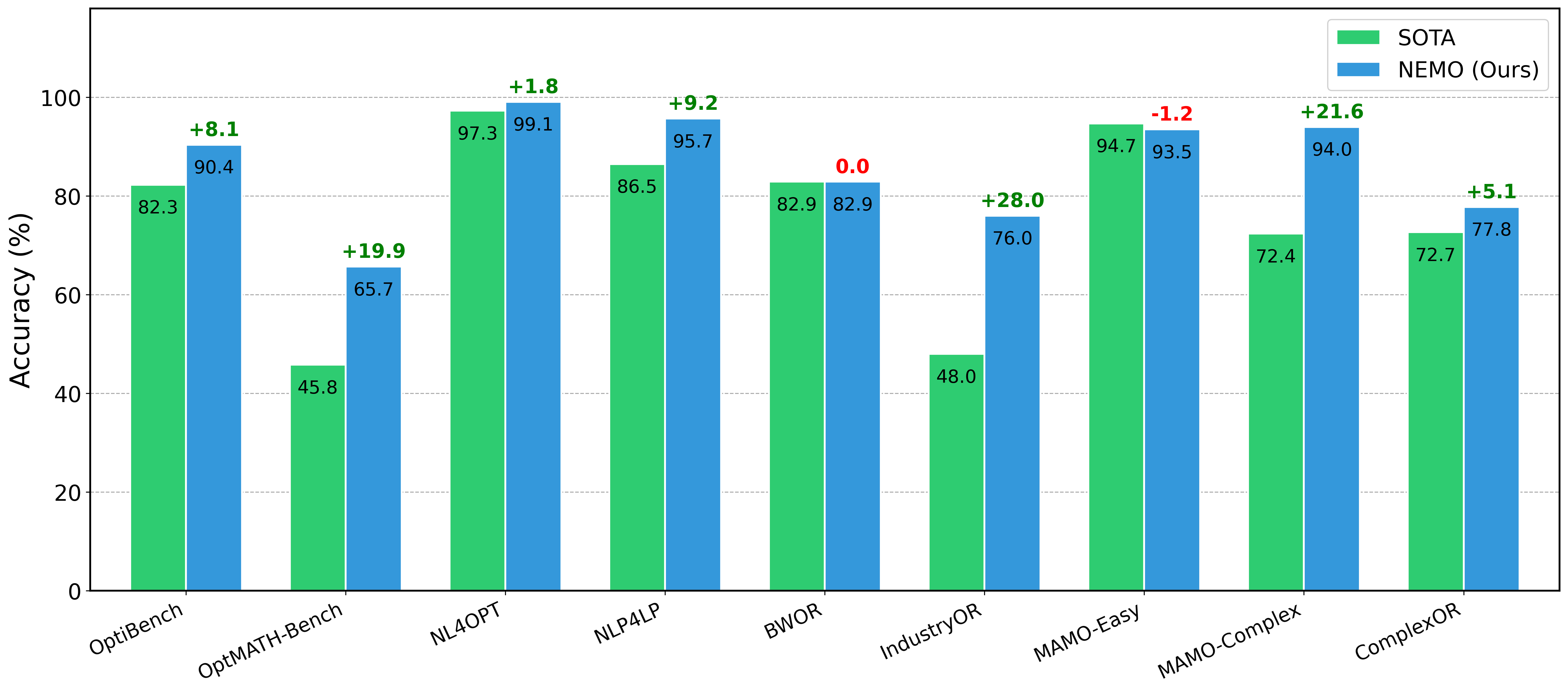}}
    \caption{Accuracy comparison between NEMO and the reported SOTA results across nine optimization benchmarks. NEMO outperforms prior SOTA on eight of nine benchmarks, with absolute gains of up to 28 percentage points. Full results are reported in Table~\ref{tab:comprehensive_results}.}
    \label{fig:sota_comparison}
  \end{center}
\end{figure}

\section{Introduction}
Optimization-based decision problems arise across a wide range of domains, including supply chain management, resource allocation, portfolio construction, and energy systems planning \cite{SINGH2012167,Saghafian03042015,SHAKOOR20161048,Cornuéjols_Peña_Tütüncü_2018,alma9924699147902466,decroix2021operating}. These problems often involve thousands of variables, complex constraints, and domain-specific structure, requiring careful formulation and expert knowledge to solve reliably. As a result, developing effective optimization solutions remains a labor-intensive process that depends on close collaboration between end-users, domain experts, and highly skilled operations research practitioners.

This development process is inherently iterative. Beyond initial formulation, optimization models must be repeatedly revised as business objectives evolve, operational constraints change, and new data becomes available. These feedback loops, spanning problem specification, solver selection, formulation, implementation, and evaluation, are costly and slow. This in turn creates a significant bottleneck that limits access to optimization-driven decision-making. Consequently, the value of optimization technologies remains largely confined to organizations with sustained access to specialized expertise. 

At a high level, this workflow consists of three recurring steps, namely, identifying the key components of the decision process (e.g., decision variables, constraints, objectives, and exogenous factors), selecting appropriate solution techniques, and formulating and implementing the corresponding optimization model. The resulting solutions are then evaluated by domain experts, often by mentally simulating system behavior and assessing feasibility and plausibility, before further refinement (a simulator-optimizer feedback loop).

Recent advances in LLMs offer a promising avenue to lower this barrier by automating parts of the optimization modeling pipeline. Prior work has explored both training-based approaches, which fine-tune LLMs for optimization tasks \cite{huang2024orlm,chen2025solver,JiangShu2025llmopt}, and agent-based frameworks that orchestrate general-purpose LLMs through specialized components \cite{xiao2024chainofexperts,optimai2025,ahmaditeshnizi2024optimus,ORLLMAgent2025}. While these methods have demonstrated encouraging progress, they suffer from fundamental limitations. Because they rely primarily on direct code generation without execution-aware validation (or ad-hoc versions of execution-based debugging), they are often brittle, frequently producing syntactically invalid or non-executable implementations. More importantly (even when relying on execution-aware debugging), they lack the sophistication and ability to instantiate the simulator–optimizer feedback loops that practitioners rely on to uncover logical inconsistencies and modeling errors, as doing so requires generating, executing and refining both simulation and optimization code iteratively and collaboratively.

In this paper, we propose a system (NEMO) that combines direct usage of LLMs with remote interaction with ACAs to enable reliable, execution-aware translation of natural-language decision descriptions into optimization models. Our design is explicitly inspired by the human-in-the-loop workflow used by optimization practitioners. By leveraging ACAs that are equipped with sandboxed execution environments, the system ensures that generated implementations are executable by construction and can be systematically validated and refined.

We evaluate NEMO in fully autonomous mode across nine established optimization benchmarks. Despite relying only on widely available and general-purpose LLMs that predate recent frontier releases, NEMO achieves state-of-the-art performance on eight benchmarks and competitive results on the remaining one, see Figure~\ref{fig:sota_comparison} and Table~\ref{tab:comprehensive_results}. These results demonstrate that execution-aware, agentic architectures can substantially improve the robustness and reliability of language-driven decision optimization.

\section{ACAs for Optimization Modeling}
A central abstraction in NEMO is remote interaction with ACAs, acting as execution-capable counterparts to LLMs. Unlike standard LLM calls that produce text-only outputs, ACAs operate within sandboxed execution environments that support code generation, execution, inspection, and iterative modification, enabling execution-aware validation.

The system interacts with an ACA through a remote interface that submits task specifications, comprising natural-language instructions, structured problem descriptions, and references to existing artifacts, and receives executable code, execution traces, and results in return. While ACA interactions are stateless at the interface level, they may reference persistent artifacts and memory managed by the system, allowing asynchronous coordination while preserving isolation and reproducibility. In our implementation, we instantiate this abstraction using \emph{OpenHands} \cite{wang2025openhands}, though the framework is agnostic to the underlying ACA platform.

\subsection{Opportunities with ACAs}
\label{sec:opportunities}
The use of ACAs as a first-class abstraction for optimization modeling introduces distinct advantages over approaches based on specialized LLMs or bespoke task-specific agents. We refer to this as a Workspace-State paradigm: agents co-inhabit a persistent sandboxed environment and interact via executable artifacts, in contrast to the message-passing paradigm of prior systems that exchange structured text through rigid I/O schemas. This design yields three key capabilities. First, the ACA acts as an architect: generated code is executable by construction, with execution, debugging, and recovery handled natively by the ACA itself, and each agent constructs a self-contained package managing dependencies, utilities, and unit tests, eliminating the need for a separate critic to interpret errors. Second, a shared workspace allows independent ACAs instantiated for different roles (e.g., simulation and optimization) to co-reside, enabling direct module imports and cross-validation without schema negotiation. Third, executable memory incorporates prior experience and exemplars directly into the ACA codebase as runnable files rather than static prompt text, allowing the agent to execute or import from them. This clean separation between high-level decision reasoning and low-level code execution underpins the asymmetric validation and coordination mechanisms described in Section~\ref{sec:asymmetric_validation}.

\subsection{Challenges with ACAs}
\label{sec:challenges}
While ACAs provide significant advantages, this paradigm introduces unique challenges that motivate our technical contributions. ACAs exhibit inherent non-determinism in both code structure and execution outcomes, manifesting as differences in variable naming, constraint formulation, solver configuration, and numerical precision. Moreover, while sandboxed execution guarantees syntactic validity, it does not ensure semantic correctness, generated code may execute successfully yet encode an incorrect formulation or violate problem constraints. Without ground-truth solutions, validating semantic correctness becomes particularly challenging. These challenges motivate the systematic mechanisms introduced in Section~\ref{sec:system_architecture}.

\section{Methodology}
\label{sec:system_architecture}
\begin{figure*}[t]
  \begin{center}
    \centerline{\includegraphics[width=0.95\linewidth]{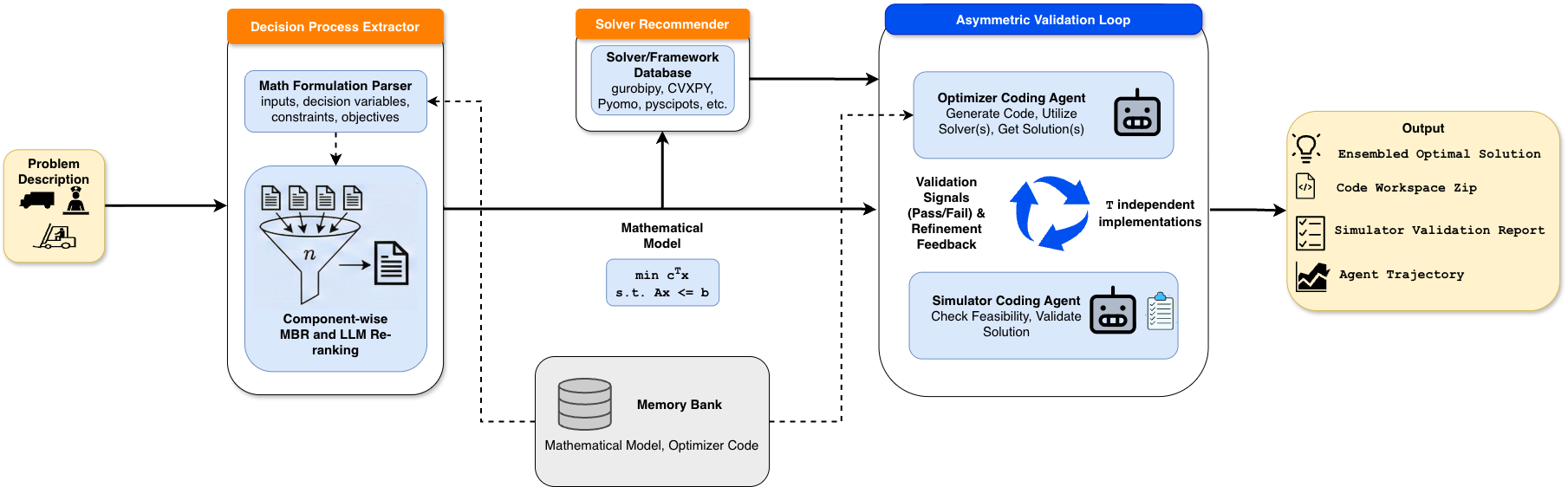}}
    \caption{
     Overview of NEMO. Natural language descriptions are translated into formal mathematical models via component-wise MBR decoding. These models drive an asymmetric validation loop between independent optimizer and simulator agents, where the simulator detects feasibility errors and guides iterative refinement. The system leverages external memory and solver recommendations to produce validated, executable optimization code.
    }
    \label{fig:system_overview}
  \end{center}
\end{figure*}

\subsection{Method Overview}

NEMO leverages the benefits of ACAs and addresses the challenges identified in Section~\ref{sec:challenges} through a coordinated multi-component architecture (Figure~\ref{fig:system_overview}) consisting of four primary modules: a decision process extractor that converts natural-language descriptions into structured representations using consensus-based decoding (Section~\ref{sec:decision_extractor}); a simulator that constructs an executable model to evaluate feasibility and objective values (Section~\ref{sec:simulator}); a solver recommender that selects appropriate optimization backends (Section~\ref{sec:solver_recommender}); and an optimizer that generates and refines executable solver code using self-consistency mechanisms (Section~\ref{sec:optimizer}).

The system exploits the asymmetry between simulation and optimization complexity through a validation loop in which the simulator serves as a fixed executable reference for validating optimizer outputs (Section~\ref{sec:asymmetric_validation}). To further improve robustness, selected modules employ diversity-aware memory retrieval for few-shot learning (Section~\ref{sec:memory}), MBR decoding to stabilize extractions (Section~\ref{sec:mbr}), and self-consistency aggregation to ensure solution reliability (Section~\ref{sec:optimizer}).

\subsection{Memory for Few-shot Learning} \label{sec:memory}
The effectiveness of in-context learning and few-shot examples for improving the performance of LLM-based systems is well established \cite{brown2020language,li2021prefix,schick2021exploiting,gpt4_arxiv2023,liu2023pre}. Motivated by this, we equip both the decision process extractor and the optimizer with access to a shared memory that enables reuse of prior problem-solving experience beyond standard prompt-based conditioning.

We construct this memory using a subset of the OptMATH \cite{lu2025optmath} training dataset, which provides diverse and structured examples of optimization problems. Each sample \( i \) in the dataset is represented as a triplet \( (D_i, I_i, C_i) \), where \( D_i \) denotes a natural-language problem description, \( I_i \) the corresponding mathematical formulation, and \( C_i \) the associated optimization code. From this dataset, we select a memory bank of 3{,}000 samples chosen to maximize coverage across 15 distinct problem types including knapsack, scheduling, routing, and facility location problems (see Appendix~\ref{app:vectorstore} for the complete taxonomy).

To enable efficient retrieval, we embed all problem descriptions \( D_i \) into a dense vector space and construct a vectorstore over these embeddings. Given a new problem description \( D \), we first retrieve a candidate pool \( \mathcal{M} \) of the top-\( N \) examples based on cosine similarity,
\[
\text{sim}(D, D_i) = \cos\big(\text{embed}(D), \text{embed}(D_i)\big),
\]
where \( \text{embed}(\cdot) \) denotes a dense embedding function. We restrict retrieval to problem descriptions, as this is the only modality available at inference time prior to formulation, and empirical similarity in this space provides sufficient signal for identifying structurally related optimization problems.

From the candidate pool \( \mathcal{M} \), we select a subset \( \mathcal{M}^* \) of \( k \) samples using a greedy strategy that balances relevance and diversity. We initialize \( \mathcal{M}^* \) with the single candidate in \( \mathcal{M} \) most similar to \( D \). We then iteratively add the candidate \( c \in \mathcal{M} \setminus \mathcal{M}^* \) that maximizes the following scoring function until \( |\mathcal{M}^*| = k \),
\[
\text{score}(c) = \text{sim}(D, c) - \lambda \cdot \frac{1}{|\mathcal{M}^*|} \sum_{m \in \mathcal{M}^*} \text{sim}(c, m).
\]
The second term penalizes redundancy by measuring the average similarity between the candidate and the examples already selected in \( \mathcal{M}^* \). This formulation keeps both similarity and diversity terms bounded in \([0,1]\), ensuring consistent behavior of the trade-off parameter \( \lambda \) across retrieval steps.

Although all candidates in \( \mathcal{M} \) are highly similar to the target problem, incorporating diversity mitigates bias toward frequently occurring patterns and guards against collapse to near-duplicate examples. To validate the impact of this parameter, we provide an ablation study over $\lambda$ in Appendix \ref{app:diversity}. Retrieved samples are used as soft guidance rather than hard constraints. Their associated formulations \( I_i \) are provided to the decision process extractor, while code artifacts \( C_i \) are supplied to the optimizer. Notably, we employ different mechanisms for incorporating retrieved examples into the decision process extractor and the optimizer, respectively; these module-specific integration strategies are described in Sections~\ref{sec:decision_extractor} and~\ref{sec:optimizer}.

\subsection{Decision Process Extractor} \label{sec:decision_extractor}

The decision process extractor is responsible for translating a natural-language description of a decision problem into a structured, machine or human interpretable representation. To this end, we leverage a carefully prompted reasoning LLM to extract the key components that define a decision process. Inspired by the decision modeling framework of \citet{pow:22}, given a natural-language description \( D \) and a set of retrieved examples \( \mathcal{M}^* \), the extractor produces a structured representation \( \mathcal{P} \) consisting of the following elements: decision variables, exogenous variables and uncertainties, state variables, transition dynamics, objective function, and constraints. In addition to the structural components, \( \mathcal{P} \) also contains inferred default values for exogenous variables and other parameters that specify the objective function and constraints, extracted from \( D \). Formally, this extraction can be expressed as \(\mathcal{E}: (D, \mathcal{M}^*) \rightarrow \mathcal{P}\).

A central challenge in using reasoning LLMs for decision process extraction is their inherent non-determinism. Even when conditioned on identical inputs, such models can produce variable outputs in terms of structure, formatting, and interpretation of extracted components. Because the decision process extractor operates at the upstream end of the system pipeline, variability at this stage can propagate to downstream modules, leading to instability in optimization formulation, execution, and our benchmarking. To mitigate this issue, we employ a variant of MBR decoding that is discussed in Section~\ref{sec:mbr}.

\subsubsection{Hybrid Component-wise MBR and LLM Re-ranking}
\label{sec:mbr}
The goal of the decision process extractor is to produce a stable and reliable structured view of different components of a decision process, without requiring manual validation or access to ground-truth formulations. To mitigate the inherent non-determinism of reasoning LLMs, we adopt a parallel extraction strategy based on MBR decoding combined with lightweight LLM-based re-ranking. The core idea is to generate multiple candidate extractions in parallel and select a representative extraction that is maximally consistent with the others, thereby reducing inconsistencies and formatting variability.

Our hybrid MBR approach consists of two stages. In the first stage, we generate \( n \) candidate extractions, conditioned on the problem description \( D \) and retrieved memory context \( \mathcal{M}^* \). Each candidate extraction \( \mathcal{P}_i \) is represented as a collection of structured components \( \{c^i_j\}_{j=1}^J \), where \( j \) indexes the component type. To quantify agreement across candidates, we compute component-wise utility scores based on semantic similarity. Each component is embedded using a dense embedding model, and similarity between components is measured via cosine similarity. For a given component type \( j \) of the \( i \)-th candidate, its similarity to the corresponding components from other candidates is defined as
\[
S(c^i_j) = \frac{1}{n-1} \sum_{\substack{k=1 \\ k \neq i}}^n \text{sim}(c^k_j, c^i_j).
\]
The overall utility score for candidate \( \mathcal{P}_i \) is computed as a weighted sum of its component utilities,
\[
U(i) = \sum_{j=1}^J w_j S(c^i_j)
\]
with fixed weights \( w_j \geq 0 \) such that \( \sum_{j=1}^J w_j = 1 \). These weights quantify the relative importance of the mathematical components of the formulation (e.g., constraints vs. variables). In our experiments, we fix the weights \( w_j \) across all candidates to reflect the relative contribution of each component type; details of the weight settings are provided in Appendix~\ref{app:hyp_config}.

Based on these utilities, we select the indices of the top-\( q \) extractions as
\[
\mathcal{I}_{\text{top-}q} = 
\operatorname*{arg\,max}_{\substack{\mathcal{I} \subseteq \{1, \ldots, n\} \\ |\mathcal{I}| = q}}
\;
\sum_{i \in \mathcal{I}} U(i).
\]
In the second stage, from this subset of extractions, a final extraction is chosen using an LLM-based logical verifier that assesses mathematical consistency, constraint completeness, and overall formulation soundness,
\[
\mathcal{P}^* = \text{LLM-Judge}(\{ \mathcal{P}_i : i \in \mathcal{I}_{\text{top-}q} \}, D).
\]
We intentionally restrict the LLM-Judge to the original problem description \( D \), rather than the full memory context, to avoid biasing the final selection toward any particular retrieved example and to ensure that the chosen extraction is logically consistent with the target problem specification. An overview of the complete pipeline is shown in Figure~\ref{fig:mbr}.

\subsection{Solver Recommender} \label{sec:solver_recommender}
Given the extraction \( \mathcal{P}^\ast \) and a set of available solvers and frameworks, \( \mathcal{SO} \), the solver recommender leverages a reasoning model to generate a ranked list of suitable solvers for solving the problem (together with certain usage and installation guidelines) as \( \mathcal{R}: (\mathcal{P}^\ast, \mathcal{SO}) \rightarrow \mathcal{SO}^* \), where \( \mathcal{SO}^* = \{(s_1, r_1, p_1), (s_2, r_2, p_2), \dots, (s_m, r_m, p_m)\} \) with \( s_i \) representing a solver, \( r_i \) being the rank of the solver (lower the better) and \( p_i \) denoting its suitability reasoning and other accompanying information.

\subsection{Simulator} \label{sec:simulator}

Given a natural-language problem description \( D \) and the extracted decision process components \( \mathcal{P}^\ast \), we construct an executable simulator that evaluates candidate decision variables against the implied process dynamics and constraints. To this end, we remotely provide instructions in a carefully constructed prompt to the ACA to generate the simulator as a self-contained Python package, defined as the mapping \(\mathcal{G}_{\text{sim}} : (D, \mathcal{P}^\ast) \rightarrow \mathcal{S}\), where \( \mathcal{S} \) denotes the resulting executable simulator.

The simulator is designed to mimic the practitioner’s internal mental model of the decision process. Given a candidate assignment to the decision variables, the coding agent orchestrates execution of \( \mathcal{S} \) and returns a structured evaluation consisting of feasibility status, detected constraint violations, and the incurred objective value. This execution-based feedback provides a concrete, model-grounded signal that is used for downstream validation and refinement.

Formally, the simulator implements a mapping
\begin{align*}
& \mathcal{S} : \mathbb{R}^{|X|} \rightarrow \{0,1\} \times (\mathbb{R} \cup \{\infty\}), \\
& \mathcal{S}(x) = (\text{feasible}(x), F_{\text{sim}}(x)),
\end{align*}
where \( x \in \mathbb{R}^{|X|} \) denotes an assignment to the decision variables and \( F_{\text{sim}}(x) \) denotes the corresponding objective value computed by the simulator (set to $\infty$ if infeasible). When infeasibility is detected, the simulator reports the violated constraints and associated diagnostic information, which is subsequently used in the asymmetric validation loop described in Section~\ref{sec:asymmetric_validation}.

\subsection{Optimizer} \label{sec:optimizer}

Analogous to the simulator, the optimizer is generated and executed through an ACA as a self-contained Python package. Given the extracted decision process components \( \mathcal{P}^\ast \), solver recommendations \( \mathcal{SO}^* \) (provided in the prompt), and retrieved code artifacts from \( \mathcal{M}^* \) (uploaded to the ACA sandbox), the ACA constructs an executable optimizer via the mapping \(\mathcal{G}_{\text{opt}} : (\mathcal{P}^\ast, \mathcal{SO}^\ast, \mathcal{M}^\ast) \rightarrow \mathcal{O}\), where \( \mathcal{O} \) denotes the resulting optimization package.

Once generated, the ACA orchestrates the execution of \( \mathcal{O} \) to solve the underlying optimization problem. This process includes invoking the selected solver, post-processing solver outputs, interpreting results, and collecting diagnostic information. The optimizer returns the optimal decision variables, solver termination status, and the corresponding objective value. To further improve the performance and robustness of the optimizer, we employ a self-consistency mechanism based on the computed decision variables, objective values, and solver status, described in Section~\ref{sec:self_consistency}.

\subsubsection{Self-Consistency for Solution Trajectories}
\label{sec:self_consistency}

We construct \( T \) optimization implementations in parallel using the ACA. Each implementation produces a candidate solution \( x_i \in \mathbb{R}^{|X|} \), along with associated solver metadata. We then select a robust solution through a hierarchical consensus procedure that aggregates solver status, objective value, and decision variables.

We first determine a consensus solver status using majority voting across the \( T \) runs. In the event of ties, we apply a lexicographic tie-breaking rule: \(\emph{Optimal} \succ \emph{Time Limit} \succ \emph{Infeasible} \succ \emph{Unbounded} \succ \emph{Error}\). This ordering favors potentially valid solutions—even if suboptimal due to time limits—over definitive failure modes.

If the consensus status is \emph{Optimal} (or \emph{Time Limit}), we further group solutions sharing this status based on their objective values, relying on numerical similarity. Two objective values \( F_{\text{opt}}(x_i) \) and \( F_{\text{opt}}(x_j) \) are considered similar if
\begin{align*}
\lvert F_{\text{opt}}(x_i) - F_{\text{opt}}(x_j) \rvert \le \text{atol} + \text{rtol} \cdot \lvert F_{\text{opt}}(x_j) \rvert,
\end{align*}
with \( \text{rtol} = 10^{-6} \) and \( \text{atol} = 10^{-9} \). The consensus objective value \( F_{\text{opt}}(x^\ast) \) is selected as the median of the largest similarity group, reducing sensitivity to floating-point noise. The final decision vector \( x^\ast \) is taken from the implementation corresponding to this median; if multiple implementations achieve the median, we select the one with the lowest solver runtime to favor efficiency. This mechanism is executed automatically by the ACA and stabilizes optimizer performance.

\subsection{Asymmetric Validation via Simulator--Optimizer Feedback}
\label{sec:asymmetric_validation}

A key technical insight underlying NEMO is the complexity gap between verification (simulation) and solving (optimization). While constructing an optimizer requires translating natural language into complex declarative mathematical constraints, constructing a simulator typically involves writing imperative Python code that directly reflects the problem logic and is empirically less prone to translation errors (Section~\ref{sec:complexity_gap}). To further ensure reliability, we instruct the ACA to generate not only the simulator \( \mathcal{S} \) but also a comprehensive suite of unit tests (implemented via \texttt{pytest}), derived from both the problem description \( D \) and the extracted formulation \( \mathcal{P}^\ast \). The simulator is used as a validation reference only if it passes these self-generated consistency checks.

NEMO leverages this validated simulator through an asymmetric cross-validation loop. Given a candidate solution \( x^\ast \) and objective value \( F_{\text{opt}}(x^\ast) \) produced by the optimizer, the simulator provides an independent execution-based evaluation:
\begin{align*}
\mathcal{S}(x^\ast) &= (\text{feasible}(x^\ast), F_{\text{sim}}(x^\ast)), \\
V(x^\ast) &=
\begin{cases}
1, & \text{if } \text{feasible}(x^\ast) = 1 \;\text{and}\; \\
   & |F_{\text{sim}}(x^\ast) - F_{\text{opt}}(x^\ast)| \le \delta, \\
0, & \text{otherwise},
\end{cases}
\end{align*}
where \(\delta = \text{atol} + \text{rtol} \cdot |F_{\text{opt}}(x^\ast)|\) is the numerical tolerance threshold. A validation outcome \( V(x^\ast) = 1 \) indicates consistency between the optimizer’s declarative formulation and the simulator’s validated imperative logic.

When validation fails (\( V(x^\ast) = 0 \)), the simulator produces a structured error report describing violated constraints or objective mismatches. This report is injected into the optimizer ACA’s context as a refinement prompt, explicitly instructing the agent to debug the optimization model against the reported failures. The optimizer then generates a revised implementation, forming a self-correcting feedback loop driven by execution artifacts rather than manual intervention.

\section{Experiments}
\subsection{Experimental Setup}

We evaluate NEMO on nine established optimization benchmarks spanning diverse problem domains and complexity levels; specifications are provided in Appendix~\ref{app:dataset_desc}. Throughout the paper, we distinguish between the Standard benchmarks (the original benchmark distributions released in prior work) and the Curated benchmarks (a curated version obtained via dataset curation and quality control; see Appendix~\ref{app:dataset_cleaning}). To ensure fair comparison with prior baselines, all results reported in the main text (Table~\ref{tab:comprehensive_results}) are evaluated on the Standard Benchmarks unless otherwise noted.

All experiments utilize a unified system configuration: OpenAI's o3 model serves as the primary reasoning LLM, while \emph{OpenHands} (powered by Claude 3.7 Sonnet) acts as the ACA. Qwen3-Embedding-8B~\cite{qwen3embedding} drives MBR and memory retrieval. To assess generalization, we avoid benchmark-specific hyperparameter tuning. Instead, system parameters (listed in Appendix~\ref{app:hyp_config}) were selected based on qualitative monitoring of a small set of development instances and applied uniformly across all tasks. This protocol emphasizes robustness and transferability over narrow optimization.

We compare NEMO against state-of-the-art agent-based frameworks, OptimAI, OptiMUS, OR-LLM-Agent, and Chain-of-Experts (CoE), and training-based methods such as ORLM, SIRL, OptMATH, and LLMOPT. 
For each baseline, we report the most recent publicly available results, taken from either the corresponding publication or the associated repository, whichever yields the strongest performance. 
Following prior work \cite{chen2025solver}, we measure accuracy by comparing the objective value produced by the system, $F_{\text{opt}}(x^\ast)$, to the ground-truth optimal objective $F(x^{\text{gt}})$; see Appendix~\ref{app:eval_criteria} for full details.

\subsection{Main Results}

\begin{table*}[t]
\centering
\tiny
\setlength{\tabcolsep}{5pt}
\begin{tabular}{l | ccccccccc | cc | cc}
\toprule
\textbf{Dataset} 
 & \multicolumn{9}{c|}{\textbf{Standard Benchmarks}} 
 & \multicolumn{2}{c|}{\textbf{Curated (LLMOPT)}} 
 & \multicolumn{2}{c}{\textbf{Curated (SIRL)}} \\
\cmidrule(lr){2-10} \cmidrule(lr){11-12} \cmidrule(lr){13-14}
 & \textbf{NEMO} & \textbf{OptimAI} 
& \textbf{OptiMUS} 
& \textbf{OR-LLM-Agent} 
& \textbf{CoE}
& \textbf{OptMATH} 
& \textbf{ORLM} 
& \textbf{LLMOPT}
& \textbf{SIRL} 
 & \textbf{NEMO} & \textbf{LLMOPT} & \textbf{NEMO} & \textbf{SIRL}  \\
\midrule
\textbf{OptiBench}      & \textbf{90.4\%} & 82.3\% & - & - & - & 66.1\% & - & 66.4\% & 67.4\% & - & - & - & - \\
\textbf{OptMATH-Bench}  & \textbf{65.7\%} & - & - & - & - & 34.7\% & - & 40.0\% & 45.8\% & - & - & - & - \\
\textbf{NL4OPT}         & \textbf{98.4\%} & - & 78.8\% & 75.9\% & 64.2\% & 95.9\% & 86.5\% & - & - & \textbf{99.1\%} & 97.3\% & \textbf{98.7\%} & 98.4\% \\
\textbf{NLP4LP}         & \textbf{81.4\%} & - & 72.0\% & - & 53.1\% & - & - & - & - & \textbf{95.7\%} & 86.5\% & - & - \\
\textbf{BWOR}           & \textbf{82.9\%} & - & - & \textbf{82.9\%} & - & - & - & - & - & - & - & - & - \\
\textbf{IndustryOR}     & \textbf{63.0\%} & - & - & 36.0\% & - & 31.0\% & 38.0\% & 44.0\% & - & - & - & \textbf{76.0\%} & 48.0\% \\
\textbf{MAMO-Easy}      & 83.4\% & - & - & 82.2\% & - & \textbf{89.9\%} & 85.2\% & - & - & 92.5\% & \textbf{95.3\%} & 93.5\% & \textbf{94.7\%} \\
\textbf{MAMO-Complex}   & 72.0\% & - & - & 51.6\% & - & 54.1\% & 44.1\% & \textbf{85.8\%} & - & - &  - & \textbf{94.0\%} & 72.4\% \\
\textbf{ComplexOR}      & \textbf{77.8\%} & - & 66.7\% & - & 38.1\% & - & - & 72.7\% & - & - & - & - & - \\
\bottomrule
\end{tabular}
\vskip 0.05in
\caption{\footnotesize Benchmark results on 9 datasets. We report accuracy on Standard benchmarks and Curated variants released by prior work (LLMOPT, SIRL). NEMO achieves strong performance across both settings, consistently improving over prior agent-based baselines and remaining competitive with training-based methods without task-specific fine-tuning. Bold denotes the best result per dataset and setting\tablefootnote{Note that OptimAI reported accuracy on NLP4LP using only the 65 LP problems, whereas we evaluate on the full dataset containing 269 LP and MILP problems.}.}
\label{tab:comprehensive_results}
\vskip -0.1in
\end{table*}

Table~\ref{tab:comprehensive_results} summarizes performance across the nine benchmarks, comparing NEMO against state-of-the-art agent-based and training-based approaches.

To ensure rigorous and fair comparison, we report results on the Standard Benchmarks as well as on the specific curated test sets released by prior works (e.g., SIRL, LLMOPT) where applicable. Overall, NEMO achieves strong and consistent performance, ranking first or tied for first on eight of the nine benchmarks under at least one evaluation setting, and outperforming prior methods by large margins on several datasets. To facilitate transparency and reproducibility, we release granular intermediate outputs from the different components of NEMO via HuggingFace\footnote{\url{https://huggingface.co/datasets/c3aiia3c/nemo-icml2026}}. We hope that this release will enable deeper analysis of system behavior and encourage further investigation and development of execution-aware approaches for language-driven optimization.
\begin{remark}
We adopt the experimental protocols reported in the prior works we benchmark against. These approaches do not report statistical significance or variability measures (e.g., multiple runs or confidence intervals), and we follow the same practice for the headline results in Table~\ref{tab:comprehensive_results}, as the ACA execution loop is computationally intensive and repeated end-to-end evaluations across all nine benchmarks would be prohibitive. To nevertheless bound run-to-run variability, we conducted 5 independent evaluation runs on BWOR and ComplexOR---the two smallest benchmarks, and thus the most susceptible to stochastic variation---yielding standard deviations of $\pm$1.3\% and $\pm$2.5\%, respectively. These values fall well within NEMO's margins over prior SOTA (up to 20 percentage points), indicating that the reported gains reflect architectural contributions rather than stochastic noise, and that the MBR and self-consistency protocols effectively stabilize ACA non-determinism.
\end{remark}

Across agent-based baselines, NEMO consistently improves upon prior coordination-based approaches. Notably, on OptiBench and OptMATH-Bench, our system achieves absolute accuracy gains of over 8 and approximately 20 percentage points, respectively, compared to the strongest reported baselines. On BWOR, our method matches the best prior result while maintaining high consistency across problem variants.

When compared to training-based methods, NEMO remains competitive or superior despite relying on general-purpose language models without domain-specific fine-tuning. Most notably, on the curated IndustryOR benchmark, NEMO outperforms SIRL by a remarkable margin of 28 percentage points, highlighting the substantial advantage of execution-aware validation and memory-based adaptation in handling complex, real-world optimization tasks.

\subsection{Ablation Study}
\label{sec:ablation}
\begin{table}[t]
\centering
\resizebox{\columnwidth}{!}{
\begin{tabular}{l|ccccc}
\toprule
\textbf{Dataset} & \textbf{NEMO} & \textbf{NEMO} & \textbf{NEMO} & \textbf{NEMO} & \textbf{NEMO} \\
& \textbf{w/o Sim} & \textbf{(Base)} & \textbf{+Mem} & \textbf{+Mem+MBR} & \textbf{+Mem+MBR+Multi} \\
\midrule
\textbf{OptMATH} & 59.6\% & 63.2\% & 63.9\% & 64.5\% & \textbf{65.7\%} \\
\textbf{BWOR} & 71.9\% & 75.6\% & 80.4\% & \textbf{82.9\%} & \textbf{82.9\%} \\
\textbf{IndustryOR} & 60.0\% & 60.0\% & 62.0\% & \textbf{63.0\%} & \textbf{63.0\%} \\
\textbf{MAMO-Complex}   & 50.9\% & 54.2\% & 61.9\% & 71.0\% & \textbf{72.0\%} \\
\textbf{NLP4LP}         & 76.0\% & 77.5\% & 79.0\% & \textbf{81.4\%} & \textbf{81.4\%} \\
\bottomrule
\end{tabular}
}
\vskip 0.05in
\caption{\footnotesize Ablation study of system components across five benchmarks. Results are reported for progressively augmented variants, starting from a variant without the simulator (NEMO w/o Sim) and incrementally adding the simulator, memory, MBR decoding, and multiple optimizer backends. Bold denotes the best result per dataset.}
\label{tab:ablation_results}
\vskip -0.15in
\end{table}
Table~\ref{tab:ablation_results} presents an ablation study examining the contribution of key system components across five benchmarks spanning combinatorial optimization (BWOR), mathematical programming (OptMATH-Bench, NLP4LP), multi-domain competition problems (MAMO-Complex), and industry-grade scenarios (IndustryOR)\footnote{ComplexOR is excluded because NEMO (Base) already saturates to 100\% accuracy on the curated dataset (see Appendix~\ref{app:dataset_cleaning}), preventing meaningful component analysis due to ceiling effects.}.

Across all benchmarks, the core system components provide consistent, incremental improvements as they are progressively layered in. The simulator contributes positively on the four non-IndustryOR benchmarks (3.7 percentage points on BWOR, 3.6 on OptMATH-Bench, 3.3 on MAMO-Complex, and 1.5 on NLP4LP), and yields no net gain on IndustryOR.

To understand the IndustryOR exception, we compare NEMO Base and NEMO w/o Sim at the instance level on the 85 evaluable problems (excluding 15 with invalid ground truths). 84\% of failures (21 of 25) are shared between the two variants, with root causes uniformly upstream of the simulator's scope: extraction errors (52\%), logic/constraint errors (43\%), and solver failures (5\%). The simulator does correct all 5 implementation-level bugs it encounters (5 of 5 detected, 4 of 5 resolved); the dominant upstream failures are instead addressed by MBR and Memory, which is why the full NEMO system still achieves 28 percentage points over prior SOTA on IndustryOR despite the simulator's zero marginal contribution in the ablation. A complementary failure-mode breakdown for the complete NEMO pipeline is provided in Figure~\ref{fig:failure} of Appendix~\ref{sec:failure_analysis}; further sensitivity analyses are in Appendix~\ref{app:ablation}.

\subsection{Complexity Asymmetry between Simulation and Optimization}
\label{sec:complexity_gap}

To validate the architectural premise that simulation is a systematically simpler task than optimization, we evaluate single-shot Pass@1 of code generation for both tasks on 100 OptiBench problems against human-verified ground truths (not agent-generated unit tests). A pass is recorded when the generated code reproduces the ground-truth objective value within numerical tolerance~$\delta$; the simulator additionally must satisfy all stated constraints upon execution. LLMs achieve 97\% Pass@1 for procedural simulators versus 87\% for declarative optimizers, a gap of 10 percentage points. This margin confirms simulation as a systematically simpler task and justifies its use as the high-confidence verification reference in the asymmetric validation loop (Section~\ref{sec:asymmetric_validation}).

\subsection{Compute-Matched Baseline and Full-Stack Simulator Contribution}
\label{sec:baselines_extended}
To verify that NEMO's gains stem from its execution-grounded architecture rather than increased inference-time compute, we construct a Best-of-5 ACA baseline using the same backend (Claude 4.5 Sonnet) with 5 independent single-pass ACA calls, selecting the best-objective solution. Despite matching the compute budget, NEMO outperforms this baseline by substantial margins (Table~\ref{tab:best_of_5}), confirming that structured execution-grounded feedback---not token volume---drives the improvements.

\begin{table}[t]
\centering
\small
\resizebox{\columnwidth}{!}{
\begin{tabular}{l|cc}
\toprule
\textbf{Method} & \textbf{BWOR} & \textbf{OptMATH-Bench} \\
\midrule
Best-of-5 ACA Baseline & 79.3\% & 62.7\% \\
\textbf{NEMO}          & \textbf{86.6\%} & \textbf{68.1\%} \\
\midrule
$\Delta$ (percentage points) & +7.3 & +5.4 \\
\bottomrule
\end{tabular}
}
\vskip 0.05in
\caption{\footnotesize Compute-matched Best-of-5 ACA baseline vs.\ NEMO (Claude 4.5 Sonnet). Both conditions use the same backend and comparable token budgets. NEMO's gains come from its execution-grounded architecture, not compute volume.}
\label{tab:best_of_5}
\vskip -0.15in
\end{table}

We additionally isolate the marginal contribution of the asymmetric validation loop within the fully assembled system. Disabling only the simulator from the complete NEMO pipeline (Memory + MBR + Multi-Optimizers) on Claude 4.5 Sonnet drops BWOR accuracy from 86.6\% to 80.5\% (a drop of 6.1 percentage points), confirming that the simulator continues to provide significant independent value even atop all other components.

\section{Related Work}
\subsection{LLM-Based Optimization Modeling}
Recent advances in LLMs have enabled significant progress toward automating optimization modeling from natural-language problem descriptions. Existing approaches broadly fall into two categories: agent-based frameworks and training-based methods.

\textbf{Agent-based frameworks} mitigate the complexity of optimization modeling by decomposing the workflow into specialized, coordinated sub-tasks. Chain-of-Experts (CoE) \cite{xiao2024chainofexperts} introduces a cooperative ecosystem where agents assume distinct reasoning roles, synchronized through iterative reflection. Similarly, pipelines like OptimAI \cite{optimai2025} and OR-LLM-Agent \cite{ORLLMAgent2025} structure the process into sequential stages, ranging from formulation to execution, often relying on coder--critic interactions to refine outputs. Similarly, OptiMUS \cite{ahmaditeshnizi2024optimus} prioritizes modularity to facilitate scalable Mixed-Integer Linear Programming (MILP) formulation. While these approaches successfully demonstrate the utility of specialization, they often necessitate intricate coordination protocols. Furthermore, their reliance on iterative critic agents can introduce fragility, as error propagation across stages remains a significant challenge when validation mechanisms are not grounded in execution.

\textbf{Training-based approaches} aim to internalize optimization knowledge directly into model parameters. ORLM \cite{huang2024orlm} performs supervised fine-tuning of open-source LLMs using synthetic instruction data generated via the OR-Instruct framework. Solver-Informed Reinforcement Learning (SIRL) \cite{chen2025solver} leverages external optimization solvers as verifiers, providing reward signals related to syntax validity, feasibility, and solution quality during training. LLMOPT \cite{JiangShu2025llmopt} combines multi-instruction supervised fine-tuning with model alignment techniques to generate structured optimization formulations and solver code. Similarly, OptMATH \cite{lu2025optmath} introduces a comprehensive instruction-tuning dataset derived from semi-structured optimization problems, enabling models to better bridge the gap between natural-language descriptions and mathematical formulations. While training-based methods can reduce hallucinations and improve consistency and performance, they require substantial computational resources and typically exhibit limited transferability to new optimization domains without additional retraining.

\subsection{Positioning of NEMO}
NEMO's Workspace-State paradigm (Section~\ref{sec:opportunities}) directly addresses the core limitations of prior message-passing systems: agents share a persistent workspace rather than exchanging structured text, enabling native execution, cross-component imports, and executable memory without schema negotiation. Building on this, the system employs an asymmetric validation loop in which an independently generated simulator serves as an executable reference for validating optimizer outputs, enabling systematic detection of logical inconsistencies and implementation errors without ground-truth formulations. In contrast to training-based methods, this design avoids domain-specific fine-tuning and adapts to new optimization domains through memory expansion rather than retraining.

\section{Limitations and Future Work}
\label{sec:limitations}

\paragraph{Formulation correctness and formal verification.}
Objective-value agreement with ground truth is a necessary but not sufficient condition for formulation correctness: two models may agree on the optimal objective while differing in their constraint sets or feasible regions. Formally verifying semantic equivalence between natural-language specifications and optimization models is an open problem in program synthesis; NEMO's asymmetric validation loop provides a practical mitigation by catching structural implementation errors through cross-paradigm execution checks, but full formal verification remains future work.

\paragraph{Computational overhead and inference-time trade-offs.}
A primary limitation of our ACA-based pipeline is computational cost. Compared to direct solver calls or single-pass LLM generation, our approach incurs additional overhead from iterative code generation, sandbox execution, and validation loops, taking 5--10 minutes per instance. Per-module latency profiling shows that approximately 55\% of wall-clock time is irreducible ACA reasoning (inherent to any agent-based system), 32\% is NEMO-specific overhead (validation loop and orchestration), and 13\% is the cold-start phase (extraction, solver recommendation, memory retrieval). While this latency is acceptable when optimizer construction is infrequent and artifacts are reused, it may be prohibitive for high-throughput applications. However, the consistent performance improvements suggest this computation represents a form of inference-time scaling, where increased reasoning effort yields higher solution quality. Future work should explore acceleration strategies including caching code templates, parallelizing independent ACA runs, and distilling recurring patterns into specialized components.

\paragraph{Learning from execution-based validation.}
Existing reinforcement learning approaches for optimization modeling rely on comparing outputs against ground-truth solutions, providing coarse, outcome-level signals limited by data availability. In contrast, our simulator–optimizer validation loop yields richer, ground-truth-free execution-based feedback by cross-checking independently generated components. This signal exposes where and how errors arise through feasibility checks, objective consistency, and structured discrepancies, rather than merely whether outputs match known answers. Leveraging this execution-grounded feedback as a learning signal represents a promising direction for improving the robustness and scalability of language-driven optimization systems.

\section{Conclusion}

We introduced NEMO, an execution-aware system for translating natural-language descriptions of decision problems into executable mathematical optimization programs using ACAs. In contrast to prior approaches based on direct LLM code generation or bespoke critic pipelines, NEMO treats ACAs as first-class primitives and leverages asymmetric validation between independently generated simulators and optimizers to systematically detect and correct modeling errors, augmented by memory-based few-shot learning, MBR decoding, and self-consistency mechanisms.

Across nine optimization benchmarks, NEMO achieves strong performance, ranking first or tied for first on eight benchmarks under at least one evaluation setting, with substantial improvements on complex real-world problems. These gains are achieved without domain-specific training, benchmark-specific tuning, or reliance on frontier models, demonstrating the effectiveness of execution-aware validation over training-intensive alternatives.

Beyond modeling, this work introduces a novel interaction paradigm with ACAs through execution-grounded workflows, integrating generation, validation, and iterative refinement, offering a robust architectural template for agentic systems in high-stakes domains requiring correctness and coordinated reasoning.

\section*{Impact Statement}

This paper presents work aimed at democratizing access to optimization modeling by lowering technical barriers for practitioners without specialized operations research training. While this could enable more efficient resource allocation across healthcare, logistics, and public services, we acknowledge several considerations. The system is designed as a decision-support tool requiring human oversight, particularly in high-stakes domains, as automated approaches can produce incorrect solutions for problems outside their training distribution. The computational overhead of our inference-time scaling approach (5-10 minutes per instance) may result in substantial energy consumption at scale. Additionally, while optimization technologies are fundamentally neutral, automated modeling could be applied to contexts raising ethical concerns, such as surveillance or algorithmic decision-making systems. We have also documented significant data quality issues in existing benchmarks (87.3\% retention rate), highlighting the need for high-quality, diverse evaluation datasets to ensure robust performance. Users should understand system limitations and maintain appropriate human validation, especially for critical applications.

\bibliography{references}
\bibliographystyle{icml2026}

\newpage
\appendix
\onecolumn

\section{Dataset}
\subsection{Dataset Description}
\label{app:dataset_desc}

We evaluate our system across nine operations research benchmark datasets. Dataset statistics are summarized in Table~\ref{tab:dataset_stats}.

\begin{table}[h]
\small
\centering
\begin{tabular}{l|c|l}
\toprule
\textbf{Dataset Name} & \textbf{\# Questions} & \textbf{Problem Types} \\
\midrule
\textbf{OptiBench} \cite{yang2025optibench} & 605 & LP, MILP, NLP \\
\textbf{OptMATH-Bench} \cite{lu2025optmath} & 166 & LP, MILP, NLP, SOCP \\
\textbf{NL4OPT} \cite{ramamonjison2023nl4opt} & 245 & LP \\
\textbf{NLP4LP} \cite{ahmaditeshnizi2024optimus} & 269 & LP, MILP \\
\textbf{BWOR} \cite{ORLLMAgent2025} & 82 & LP, MILP \\
\textbf{IndustryOR} \cite{huang2024orlm} & 100 & LP, MILP \\
\textbf{MAMO-Easy} \cite{huang2024mamo} & 652 & LP, MILP \\
\textbf{MAMO-Complex} \cite{huang2024mamo} & 211 & LP, MILP, NLP \\
\textbf{ComplexOR} \cite{xiao2024chainofexperts} & 18 & LP, MILP \\
\bottomrule
\end{tabular}
\vskip 0.1in
\caption{Comparison of optimization benchmark datasets by size and problem type.}
\label{tab:dataset_stats}
\vskip -0.1in
\end{table}

\textbf{OptiBench}: A collection of 605 optimization word problems sourced from university textbooks and open-source solver repositories. It spans a diverse range of mathematical formulations, including Linear Programming (LP), Mixed-Integer Linear Programming (MILP), and Non-Linear Programming (NLP).

\textbf{OptMATH-Bench}: Contains 166 challenging semi-structured instances derived from advanced mathematics competitions. The dataset is characterized by extended natural-language contexts and complex constraints covering Linear Programming (LP), Mixed-Integer Linear Programming (MILP), Non-Linear Programming (NLP), and Second-Order Cone Programming (SOCP).

\textbf{NL4OPT}: Comprises 245 Linear Programming (LP) problems synthetically generated for the NeurIPS 2022 competition. It focuses on the precise translation of natural-language descriptions into canonical linear constraints and objective functions.

\textbf{NLP4LP}: A dataset of 269 problems featuring long, intricate descriptions adapted from standard optimization libraries. While primarily focused on Linear Programming (LP), harder subsets include Mixed-Integer Linear Programming (MILP) instances that test extraction from dense technical specifications.

\textbf{BWOR}: Consists of 82 business-oriented problems sourced from classic Operations Research textbooks. These problems represent standard reasoning tasks involving Linear Programming (LP) and Mixed-Integer Linear Programming (MILP) formulations applied to typical business scenarios.

\textbf{IndustryOR}: An industrial benchmark containing 100 problems derived from real-world case studies in manufacturing, supply chain logistics, and finance. It focuses on practical applications requiring models to handle constraints common in industrial Linear Programming (LP) and Mixed-Integer Linear Programming (MILP) settings.

\textbf{MAMO-Easy}: A subset of the MAMO benchmark containing 652 problems collected from mathematical modeling competitions. These instances focus on fundamental algebraic and Linear Programming (LP) tasks suitable for evaluating basic solver capabilities.

\textbf{MAMO-Complex}: The difficult subset of the MAMO benchmark (211 problems), also sourced from modeling competitions. These instances involve intricate dependencies and often require advanced Linear, Mixed-Integer, or Non-Linear Programming (LP/MILP/NLP) formulations and multi-step reasoning.

\textbf{ComplexOR}: A small but highly challenging set of 18 problems involving complex Linear Programming (LP) and Mixed-Integer Linear Programming (MILP) scenarios. These expert-crafted instances are designed to stress advanced reasoning under intricate constraint dependencies.

\subsection{Dataset Curation and Quality Control}
\label{app:dataset_cleaning}

\subsubsection{Curation Methodology}
We applied a systematic three-stage pipeline: (1) automated validation to detect malformed problems, (2) manual inspection of edge cases, and (3) exclusion based on rigorous predefined criteria. To ensure consistency, all excluded problems were independently reviewed by at least two domain experts.

\subsubsection{Exclusion Criteria}
Problems were excluded if they exhibited the following issues:
\begin{itemize}
    \item \textbf{Malformed Problem Statements:} Descriptions that were incomplete or ambiguous, particularly those lacking necessary constraints, clear objective functions, or defined decision variables.
    \item \textbf{Invalid Reference Solutions:} Ground-truth solutions that were mathematically infeasible, violated explicit constraints (e.g., exceeding budget or capacity limits), or contained numerical anomalies (e.g., arbitrarily large constants like $10^{35}$ used as proxies for infinity).
\end{itemize}

\subsubsection{Dataset Statistics}

\begin{table}[h]
    \centering
    \small
    \begin{tabular}{l|c|c|c|c}
        \toprule
        \textbf{Dataset} & \textbf{Original} & \textbf{Excluded} & \textbf{Final} & \textbf{Retention} \\
        \midrule
        OptiBench & 605 & 54 & 551 & 91.1\% \\
        OptMATH-Bench & 166 & 44 & 122 & 73.5\% \\
        NL4OPT & 245 & 3 & 242 & 98.8\% \\
        NLP4LP & 269 & 22 & 247 & 91.8\% \\
        BWOR & 82 & 10 & 72 & 87.8\% \\
        IndustryOR & 100 & 15 & 85 & 85.0\% \\
        MAMO-Easy & 652 & 89 & 563 & 86.3\% \\
        MAMO-Complex & 211 & 58 & 153 & 72.5\% \\
        ComplexOR & 18 & 4 & 14 & 77.8\% \\
        \midrule
        \textbf{Total} & \textbf{2,348} & \textbf{299} & \textbf{2,049} & \textbf{87.3\%} \\
        \bottomrule
    \end{tabular}
    \vskip 0.1in
    \caption{Curation results across all benchmark datasets. Overall retention rate: 87.3\%.}
    \label{tab:cleaning_stats}
    \vskip -0.1in
\end{table}

\begin{table}[h]
    \centering
    \small
    \begin{tabular}{l|c|c}
        \toprule
        \textbf{Exclusion Category} & \textbf{Count} & \textbf{Percentage} \\
        \midrule
        Malformed problem statements & 83 & 27.8\% \\
        Ground-truth issues & 216 & 72.2\% \\
        \midrule
        \textbf{Total} & \textbf{299} & \textbf{100\%} \\
        \bottomrule
    \end{tabular}
    \vskip 0.1in
    \caption{Distribution of exclusion reasons across all datasets.}
    \label{tab:exclusion_breakdown}
    \vskip -0.1in
\end{table}

\subsubsection{Curation Sensitivity}
\label{app:curation_sensitivity}
Curation was strictly \emph{error-blind}: instances were excluded solely on the grounds of mathematical invalidity (infeasible ground truths, malformed problem statements), with no reference to system outputs or per-method performance. The two exclusion categories---malformed problem statements (27.8\%) and ground-truth issues (72.2\%, Table~\ref{tab:exclusion_breakdown})---are identifiable independently of any solver or model output.

To verify that headline results do not depend on curation, Table~\ref{tab:comprehensive_results} (main body) reports all primary comparisons on the Standard (uncurated) benchmarks. NEMO ranks first or tied for first on eight of nine benchmarks in this setting, confirming that the gains are not an artifact of the curation process. The curated benchmarks (Tables~\ref{tab:cleaning_stats}--\ref{tab:ablation_results_clean_dataset}) are provided for transparency and to enable fair comparison with prior works that evaluated on their own curated subsets; they consistently show the same relative ordering and larger absolute margins (as invalid ground truths are removed), but are not required for any of the paper's primary claims.

\subsubsection{Representative Exclusion Examples}

Below are two representative examples of rejected instances that were filtered out during this process.
\FloatBarrier
\begin{resultbox}{Malformed Statement (MAMO-Easy)} PROBLEM: A manufacturing company produces two types of products, X and Y. The production cost for each unit of product X is $5000, and for product Y, it's $3000. The total units produced for both products combined cannot exceed 1000 due to capacity constraints. To meet the market demand, the combined units produced, calculated as 3 times the units of product X plus 2 times the units of product Y, must be at least 2000. Additionally, the difference between the units produced for product X and Y should be at least 150 to maintain a balanced portfolio. Given that all variables are integers due to the indivisible nature of physical goods, what is the minimum total production cost in dollars (rounded to nearest dollar) required to fulfill these requirements?

PROVIDED GROUND-TRUTH (GT): 3,230,000

ISSUE: The ground truth of 3,230,000 is a result because it indicates that the solver interpreted the Balance Constraint (|x - y| being greater or equal to 150) as the more restrictive x - y being greater than 150 (Case 1), rather than allowing for more of the cheaper product (Case 2), which results in 3,000,000 as the solution.
\end{resultbox}
\vspace{1cm}
\FloatBarrier
\begin{resultbox}{Unbounded Ground-Truth (IndustryOR)}
PROBLEM: A company must allocate a $10,000 weekly budget across three advertising platforms to maximize viewer reach: Z-tube ($1,000/ad, 400,000 viewers), Soorchle ($200/ad, 5,000 viewers), and Wassa ($100/ad, 3,000 viewers). Constraints require: (1) at most 15 ads on Soorchle, (2) at most one-third of total ads on Wassa, and (3) at least 5% of total ads on Z-tube.

PROVIDED GROUND-TRUTH (GT):
{
    "variables": {
        "NumberAdsZTube": -0.0,
        "NumberAdsSoorchle": -0.0,
        "NumberAdsWassa": -0.0,
        "xZ": 1e+30,
        "xS": 1e+30,
        "xW": 1e+30
    },
    "objectives": 4.08e+35
}

ISSUE: The ground truth solution is unbounded whereas the there exists feasible solution to this problem.
\end{resultbox}

\section{Ablation Studies}
\label{app:ablation}
\subsection{Performance on Curated Benchmarks} 
\begin{table}[h]
\centering
\small
\begin{tabular}{l|ccccc}
\toprule
\textbf{Dataset} & \textbf{NEMO} & \textbf{NEMO} & \textbf{NEMO} & \textbf{NEMO} & \textbf{NEMO} \\
& \textbf{-Sim} & \textbf{(Base)} & \textbf{+Mem} & \textbf{+Mem+MBR} & \textbf{+Mem+MBR+Multi} \\
\midrule
\textbf{OptMATH-Bench} & 81.1\% & 86.0\% & 86.8\% & 87.7\% & \textbf{89.3\%} \\
\textbf{BWOR} & 81.9\% & 86.1\% & 91.7\% & \textbf{94.3\%} & \textbf{94.3\%} \\
\textbf{IndustryOR} & 71.4\% & 71.4\% & 73.8\% & \textbf{75.0\%} & \textbf{75.0\%} \\
\bottomrule
\end{tabular}
\vskip 0.1in
\caption{Ablation study showing performance across NEMO variants on the Curated benchmarks. Bold indicates best performance.}
\label{tab:ablation_results_clean_dataset}
\vskip -0.1in
\end{table}
Table~\ref{tab:ablation_results_clean_dataset} details how each system component contributes to performance on the curated datasets. We utilize these verified benchmarks to measure the model's true reasoning capabilities, isolating them from the noise caused by malformed or incorrect problems present in the original distributions.

The results demonstrate a clear, consistent trajectory of improvement as components are layered in. For instance, on the BWOR benchmark, the addition of Memory and MBR decoding steadily raises accuracy from 86.1\% (Base) to 94.3\%. Overall, these findings confirm that the system's design elements function synergistically to enhance robustness. Furthermore, the significantly higher absolute scores compared to the standard benchmarks suggest that the performance gaps observed in Table~\ref{tab:comprehensive_results} are largely attributable to data quality issues in the original sources rather than intrinsic limitations of the model.

\subsection{Simulator-Optimizer Feedback Loop Analysis}
\begin{table}[t]
\centering
\small
\begin{tabular}{l|c|c|c|c}
\toprule
\textbf{Dataset} 
& \textbf{Total} 
& \textbf{Multi-Attempt} 
& \textbf{Resolved} 
& \textbf{Success Rate} \\
& \textbf{Problems}
& \textbf{Triggered}
& \textbf{Correctly}
& \textbf{(Multi-Attempt)} \\
\midrule
\textbf{OptMATH-Bench}   & 166 & 8 & 5 & 62.5\% \\
\textbf{BWOR}            & 82  & 2 & 2 & 100\% \\
\textbf{IndustryOR}      & 100 & 5 & 4 & 80\% \\
\bottomrule
\end{tabular}
\vskip 0.1in
\caption{Simulator-optimizer feedback loop activation and effectiveness across benchmarks. \textbf{Multi-Attempt Triggered} indicates problems requiring more than one iteration through the feedback loop. \textbf{Resolved Correctly} shows how many of these were ultimately solved. \textbf{Success Rate} measures the percentage of multi-attempt problems that were resolved correctly, demonstrating the effectiveness of the iterative refinement process. }
\label{tab:simulator_feedback_analysis}
\vskip -0.1in
\end{table}
Table~\ref{tab:simulator_feedback_analysis} quantifies the simulator-optimizer feedback loop activation across three benchmarks. The feedback mechanism was triggered in only 1.2-5\% of problems, indicating high initial solution quality. However, when activated, it achieved strong correction rates of 62.5-100\%, successfully resolving 11 out of 15 initially incorrect solutions. This validates the simulator's effectiveness both in detecting errors and guiding iterative refinement toward correct solutions.

\subsection{Base Model in \emph{OpenHands} (Claude 4.5 vs. Claude 3.7)}
To understand how much our system benefits from stronger underlying models, we compared performance using Claude 3.7 Sonnet versus the more capable Claude 4.5 Sonnet. Table~\ref{tab:base_model} details these results on both Standard and Curated benchmarks.

We observe that upgrading the base model yields consistent gains across all datasets. For example, on the Curated BWOR benchmark, accuracy rises to a near-perfect 98.6\%. Importantly, these improvements occur in both the Standard and Curated settings. This confirms that NEMO scales effectively with better base models and that the performance boost from our agentic framework is complementary to advancements in the underlying LLM.
\FloatBarrier

\begin{table}[h]
    \centering
    \small
    \begin{tabular}{l|cc|cc}
        \toprule
           \multirow{2}{*}{\textbf{Dataset}} & \multicolumn{2}{c|}{\textbf{Standard Benchmarks}} & \multicolumn{2}{c}{\textbf{Curated Benchmarks}} \\
        \cmidrule(lr){2-3} \cmidrule(lr){4-5}
       & \textbf{Sonnet 3.7} & \textbf{Sonnet 4.5} & \textbf{Sonnet 3.7} & \textbf{Sonnet 4.5} \\
        \midrule
        \textbf{OptMATH-Bench} & 65.7\% & \textbf{68.1\%} & 89.3\% & \textbf{92.6\%} \\
        \textbf{BWOR} & 82.9\% & \textbf{86.6\%} & 94.3\% & \textbf{98.6\%} \\
        \textbf{IndustryOR} & 63.0\% & \textbf{65.0\%} & 75.0\% & \textbf{76.4\%} \\
        \bottomrule
    \end{tabular}
    \vskip 0.1in
    \caption{Impact of base model selection. We benchmark performance on both Standard and Curated datasets. Across both data regimes, upgrading the base model from Claude 3.7 to Claude 4.5 yields significant performance gains. Bold indicates best performance.}
    \label{tab:base_model}
    \vskip -0.1in
\end{table}

\subsection{Vectorstore Analysis}
\label{app:vectorstore}
\begin{table}[h]
\centering
\small
\renewcommand{\arraystretch}{1.2} 
\begin{tabular}{lp{8cm}} 
\toprule
\textbf{Problem Type} & \textbf{Description} \\
\midrule
Knapsack & Select items to maximize value under capacity constraints \\
Assignment & One-to-one assignment of tasks, resources, or entities \\
Scheduling & Arrange timing and sequence of activities, tasks, or jobs \\
Transportation & Optimize shipment from sources to destinations \\
Facility Location & Decide where to open facilities to serve customers \\
Network Flow & Optimize resource flow in networks \\
TSP & Find shortest path visiting all nodes once \\
Vehicle Routing & Optimize delivery routes for multiple vehicles \\
Resource Allocation & Allocate limited resources among activities \\
Production Planning & Optimize production quantities and inventory \\
Inventory Management & Optimize inventory levels and ordering \\
Cutting Stock & Minimize material waste in cutting \\
Bin Packing & Pack items into minimum containers \\
Linear Programming & General linear optimization \\
Miscellaneous & Hybrid or uncategorized problems \\
\bottomrule
\end{tabular}
\vskip 0.1in
\caption{Taxonomy of problem types stored within the memory bank.}
\label{tab:problem_types}
\vskip -0.1in
\end{table}

Our memory bank contains 3,000 optimization problems spanning 15 distinct categories. Table~\ref{tab:problem_types} provides the taxonomy and definitions for these problem types. 

\begin{figure}[t]
\vskip 0.2in
\centering
\includegraphics[width=1.0\linewidth]{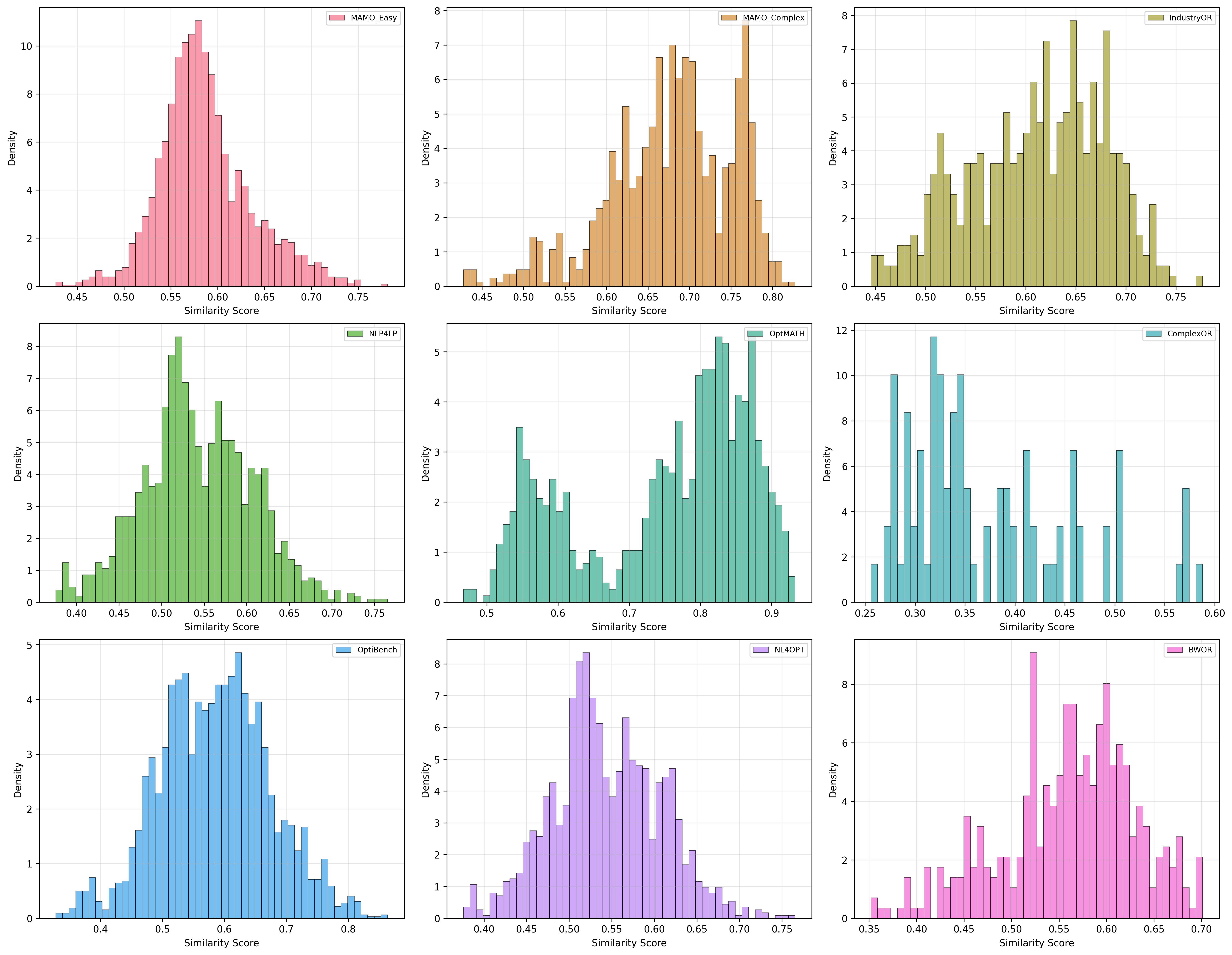}
\caption{Distribution of top-5 similarity scores for nine evaluation benchmarks against the OptMATH training set. The distributions indicate a healthy semantic gap between the test queries and the stored training samples. The absence of high-density peaks near 1.0 confirms that no significant data leakage occurs, even when retrieving for the domain-adjacent OptMATH-Bench (center panel).}
\label{fig:sim_dist}
\end{figure}

To investigate potential data leakage and the robustness of our retrieval mechanism, we populated the vectorstore with 3,000 training samples from the OptMATH-train dataset. For each evaluation problem, we retrieved the top-5 relevant samples and analyzed the resulting similarity score distributions across all benchmarks. As detailed in Figure~\ref{fig:sim_dist}, the absence of similarity scores approaching 1.0 confirms that while the retriever identifies semantically relevant structures, it does not encounter exact duplicates or leaked test data, thereby ensuring the integrity of the evaluation.

\subsubsection{Impact of Diversity Parameter $\lambda$}
Table~\ref{tab:ablation_diversity} evaluates the impact of the diversity penalty $\lambda$ in our memory retrieval scoring function. We compare a pure relevance-based strategy ($\lambda = 0.0$) against our diversity-aware approach ($\lambda = 0.5$). The results indicate that encouraging diversity is important for retrieving effective few-shot examples. On the BWOR dataset, introducing diversity improves accuracy by 13.4 percentage points, while IndustryOR exhibits an 11 percentage point gain. These findings suggest that simply retrieving the most similar examples often leads to redundant context, whereas enforcing $\lambda > 0$ promotes a broader and more representative set of problem-solving patterns, improving generalization.

\label{app:diversity}
\begin{table}[h]
\centering
\small
\begin{tabular}{l|ccccc}
\toprule
\textbf{Dataset} & \multicolumn{5}{c}{\textbf{Diversity Parameter } $\lambda$} \\
\cmidrule(lr){2-3}
& $\lambda = 0.0$ & $\lambda = 0.5$ \\
\midrule
\textbf{OptMATH-Bench} & 60.0\% & \textbf{64.4\%}  \\
\textbf{BWOR} & 69.5\% & \textbf{82.9\%}  \\
\textbf{IndustryOR} & 52.0\% & \textbf{63.0\%} \\
\bottomrule
\end{tabular}
\vskip 0.1in
\caption{Ablation study on diversity parameter $\lambda$ in the memory retrieval scoring function. $\lambda = 0$ prioritizes pure relevance while $\lambda > 0$ introduces diversity penalty. Bold indicates best performance for each dataset.}
\label{tab:ablation_diversity}
\vskip -0.1in
\end{table}
\FloatBarrier

\subsection{Consistency of MBR-Based Re-ranking}
\begin{figure}[h]
\vskip 0.2in
\centering
\includegraphics[width=0.8\linewidth]{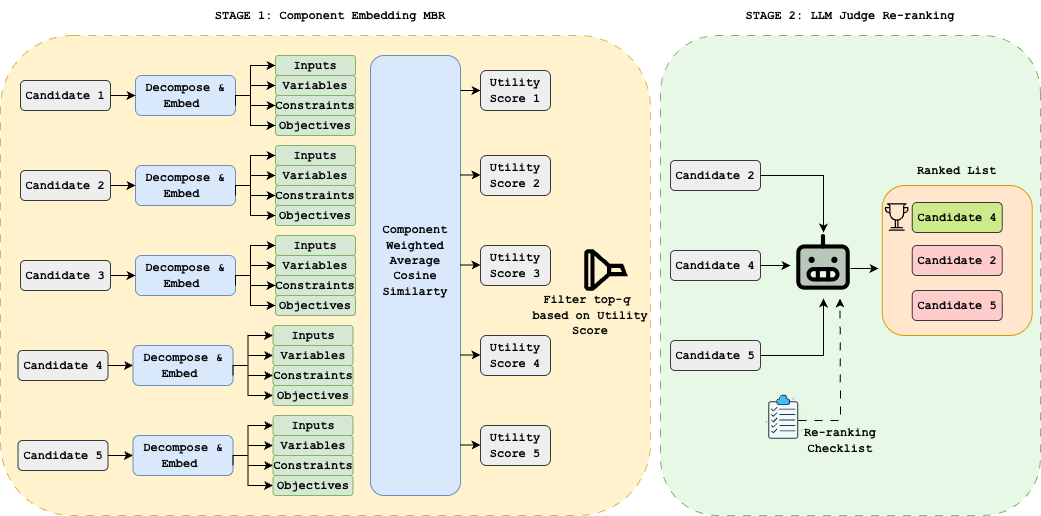}
\caption{Hybrid component-wise MBR and LLM re-ranking pipeline. A fast embedding-based filter removes weak candidates early, allowing stronger reasoning models to be reserved for final top-$q$ re-ranking, where semantic similarity is replaced by logical verification to select mathematically consistent extractions.}
\label{fig:mbr}
\end{figure}

\begin{figure}[h]
\centering
\begin{minipage}[c]{0.45\linewidth}
  \centering
  \tiny
  \includegraphics[width=\linewidth]{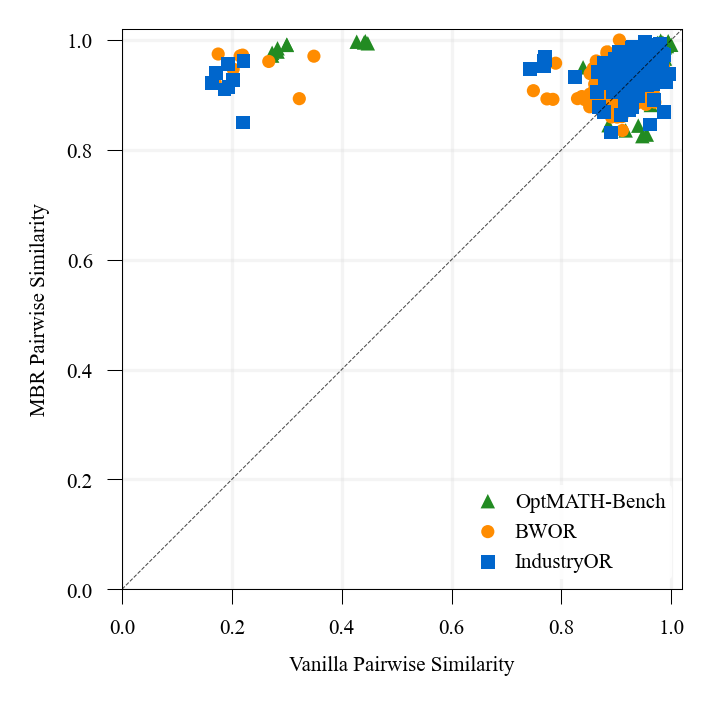}
  \captionof{figure}{Scatter plot of raw pairwise similarity scores comparing vanilla sampling and MBR decoding. The dashed line indicates parity.}
  \label{fig:mbr_scatter}
\end{minipage}
\hfill
\begin{minipage}[c]{0.45\linewidth}
  \centering
  \small
  \begin{tabular}{l|cc|cc}
    \toprule
    \textbf{Dataset} & \multicolumn{2}{c|}{\emph{Consistency} $\uparrow$} & \multicolumn{2}{c}{\emph{Stability} $\downarrow$} \\
    \cmidrule(lr){2-3} \cmidrule(lr){4-5}
     & \textbf{Vanilla} & \textbf{MBR} & \textbf{Vanilla} & \textbf{MBR} \\
    \midrule
    \textbf{OptMATH-Bench} & 0.915 & \textbf{0.960} & 0.048 & \textbf{0.015} \\
    \textbf{BWOR} & 0.895 & \textbf{0.935} & 0.057 & \textbf{0.023} \\
    \textbf{IndustryOR} & 0.899 & \textbf{0.944} & 0.059 & \textbf{0.019} \\
    \bottomrule
  \end{tabular}
  \vskip 0.1in
  \captionof{table}{Extraction variability analysis. \emph{Consistency} denotes mean pairwise similarity ($\uparrow$), while \emph{Stability} denotes intra-sample standard deviation ($\downarrow$).}
  \label{tab:mbr_stability}
\end{minipage}
\end{figure}

To reduce the stochasticity of LLM outputs, we employ MBR decoding. Figure~\ref{fig:mbr} illustrates the hybrid pipeline: a fast embedding-based filter first removes weak candidates, allowing a stronger reasoning model to focus on a small set of promising solutions. We evaluate this approach using the scatter plot in Figure~\ref{fig:mbr_scatter} and the quantitative results in Table~\ref{tab:mbr_stability}.

The scatter plot reveals two distinct behaviors. First, most points cluster in the top-right quadrant, indicating that when the model is already confident, MBR agrees with standard sampling and preserves high-quality solutions. Second, a notable set of points appears in the top-left quadrant, corresponding to cases where vanilla sampling produces inconsistent outputs, but MBR successfully identifies a consensus solution.

Table~\ref{tab:mbr_stability} quantifies this stabilizing effect. By systematically filtering inconsistent outliers, MBR substantially reduces variance across extractions. For example, on OptMATH-Bench, stability improves by approximately a factor of three, indicating that MBR effectively mitigates random failures that arise under standard sampling.

\subsection{Solver Recommender Ablation}
\label{app:solver_ablation}
To isolate the contribution of the solver recommender, we run a Gurobi-only ablation on the 133 nonlinear-notable problems in OptiBench---the subset where solver selection most critically affects outcomes. Replacing the recommender with a fixed Gurobi backend drops accuracy from 84.2\% to 76.7\% (a drop of 7.5 percentage points), confirming that matching solvers to problem structure, particularly for non-linear constraints, is a significant driver of NEMO's performance on this subset.

\section{Experiment Configuration}
\subsection{Hyperparameter Configuration}
\label{app:hyp_config}
Table~\ref{tab:hyperparameters} summarizes the global hyperparameter configuration used across all experiments. 
To ensure a consistent and reproducible evaluation, we fix a single set of parameters across all the benchmark datasets. 
\begin{table}[h]
\centering
\small
\renewcommand{\arraystretch}{1.2}
\begin{tabular}{@{} l l p{6.2cm} c @{}}
\toprule
\textbf{Category} & \textbf{Hyperparameter} & \textbf{Description} & \textbf{Value} \\
\midrule

\multirow{4}{*}{\textbf{Models}} 
 & Reasoning LLM & Primary reasoning engine for NEMO & OpenAI o3 \\
 & ACA Backend & Base model for the OpenHands agent & Claude 3.7 Sonnet \\
 & Embedding Model & Model for MBR and memory retrieval & Qwen3-Embedding-8B \\
 & Batch Size & Number of instances per batch & 5 \\
\midrule

\multirow{4}{*}{\textbf{Retrieval}} 
 & Similarity Threshold & Minimum cosine similarity score & 0.6 \\
 & Memory Pool Size ($|\mathcal{M}|$) & Number of candidates initially retrieved & 9 \\
 & Top-$k$ Retrieved ($k$) & Number of examples selected for context & 3 \\
 & Diversity ($\lambda$) & Balance between relevance and diversity & 0.5 \\
\midrule

\multirow{6}{*}{\textbf{Extractor}} 
 & Candidate Pool ($n$) & Total candidates generated & 5 \\
 & Top-$q$ Extractions & Candidates forwarded to LLM judge re-ranker & 3 \\
 & Constraint Weight & Importance weight for constraints component & 0.6 \\
 & Decision Variable Weight & Importance weight for decision variables & 0.2 \\
 & Objective Weight & Importance weight for objective function & 0.1 \\
 & Input Weight & Importance weight for input parameters & 0.1 \\
\midrule

\multirow{3}{*}{\textbf{Optimizer}} 
 & Optimizer Implementations ($T$) & Number of code implementations generated & 3 \\
 & Maximum Validation Loops & Maximum optimizer validation iterations & 3 \\
\bottomrule
\end{tabular}
\vskip 0.1in
\caption{Hyperparameter configuration. The \textbf{Category} column groups settings by module. MBR component weights (\textbf{Extractor}) are normalized to sum to 1.0.}
\vskip -0.1in
\label{tab:hyperparameters}
\end{table}

\subsection{Evaluation Criteria}
\label{app:eval_criteria}
We evaluate solution accuracy by comparing the generated objective value \( F_{\text{opt}}(x^\ast) \) against the ground-truth optimal objective \( F(x^{\text{gt}}) \). Following prior work \cite{chen2025solver}, a solution is classified as correct if it satisfies the relative error criterion
\[
\frac{\lvert F_{\text{opt}}(x^\ast) - F(x^{\text{gt}}) \rvert}{\lvert F(x^{\text{gt}}) \rvert + \epsilon} < 10^{-6},
\]
where \( x^{\text{gt}} \) denotes the ground-truth optimal solution and \( \epsilon \) is a small numerical stability constant, e.g., $\epsilon=10^{-8}$. 

In addition to satisfying this numerical threshold, we classify solutions as correct under the following well-defined exceptional cases, which arise from common ambiguities in benchmark formulations:

\begin{enumerate}
    \item \textbf{Relaxation Mismatch.}  
    The natural-language problem description implies discrete decision variables (e.g., counts of physical items), while the benchmark ground-truth is derived from a continuous LP relaxation. In such cases, solutions consistent with the relaxed formulation are considered correct.
    \item \textbf{Verified Infeasibility.}  
    The benchmark ground-truth indicates that the problem is infeasible, and the proposed solution independently proves infeasibility through execution-based validation.
    \item \textbf{Equivalent Formulations.}  
    The generated decision vector \( x^\ast \) is equivalent to \( x^{\text{gt}} \), but the reported objective values differ due to alternative scaling or units of measurement (e.g., total cost reported in USD versus thousands of USD).
\end{enumerate}

\section{Failure Modes Analysis}
\label{sec:failure_analysis}
\begin{figure}[h]
\centering
\includegraphics[width=0.6\columnwidth]{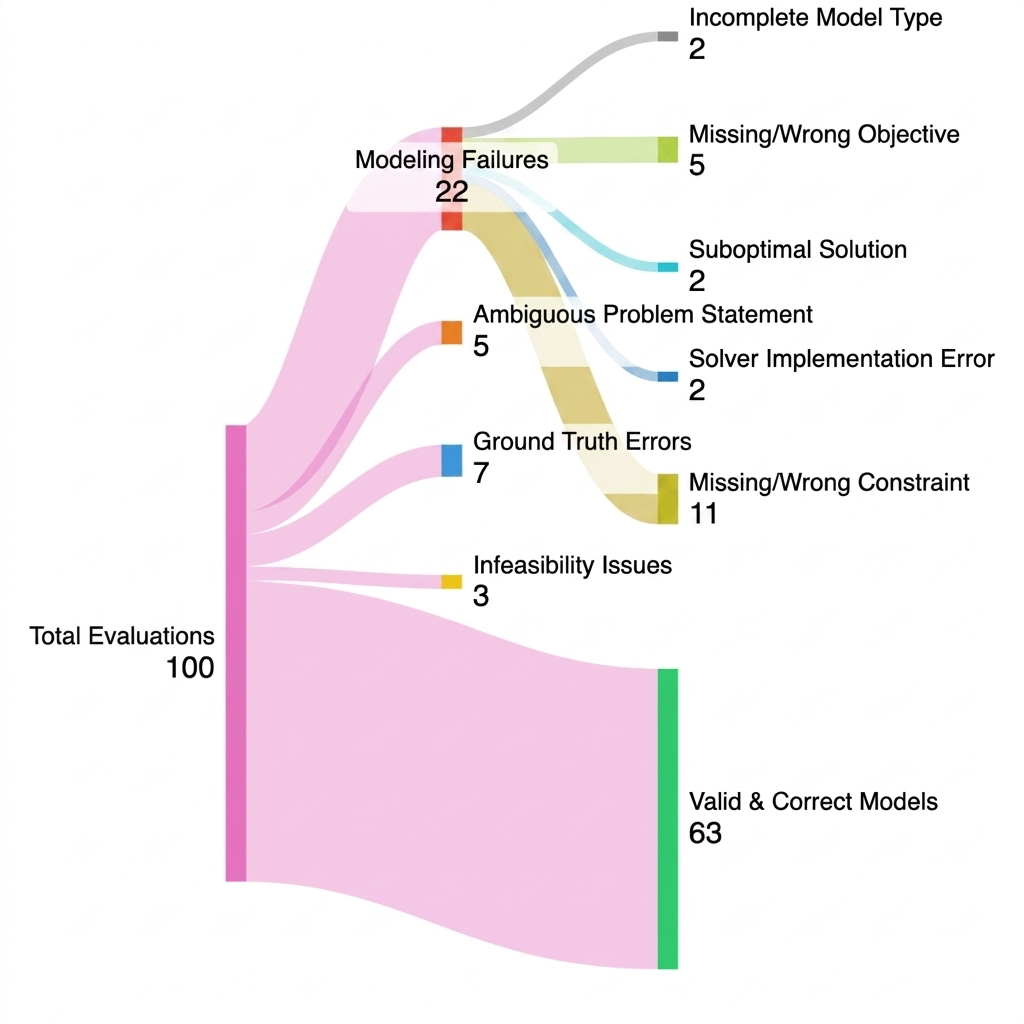}
\caption{The distribution of failure modes in NEMO's optimization pipeline on the IndustryOR benchmark. The flow transitions from the total problem set into valid models and four primary categories of failure.}
\label{fig:failure}
\end{figure}

Through a granular analysis of 100 benchmark problems from IndustryOR, we identify that NEMO achieves a 63\% success rate in generating valid and correct models. However, the remaining 37\% of cases reveal critical failure modes categorized into modeling logic, external benchmark inconsistencies, and feasibility constraints. As shown in Figure~\ref{fig:failure}, the primary bottleneck is \textbf{Wrong/Missing Constraints}, which accounts for 50\% of all modeling-related errors. This indicates that while the system often identifies the correct objective, it may overlook the physical or logical boundaries inherent in complex industrial scenarios. Additionally, 12\% of failures are attributed to \textbf{Upstream Inconsistencies} (Malformed Problem Statements or Incorrect Ground Truths), where the model's output is penalized by artifacts within the benchmark data itself rather than logical derivation errors.
\FloatBarrier

\section{Module-Specific Prompts \& End-to-End Execution Examples}
\label{app:prompts}
In this section, we provide the system prompts used in our framework. While we shortened some text for brevity, all critical logic and rules are included. We also provide examples of the results generated by the system for a MAMO-Complex minimum cost network flow problem.
\subsection{Decision Process Extractor}
\subsubsection{Prompts}
\FloatBarrier
\begin{promptbox}{Decision Process Extractor Prompt}
Extract a complete mathematical specification of an optimization problem from a natural-language description.

Core logic:
- Interpret the problem strictly from problem_description.
- Extract all explicitly stated data, parameters, variables, objectives, and constraints without omission.
- Do not summarize, truncate, or infer unstated information.
- Represent the problem using structured components:
  * inputs and parameters
  * exogenous variables and uncertainties
  * decision and state variables
  * objective function
  * constraints
  * transition function (if applicable)

Type inference logic:
- Determine variable types using a strict priority:
  1) explicit textual indicators,
  2) cost unit semantics (per-item vs per-measure),
  3) naming semantics (discrete objects vs divisible quantities).
- Do not guess or add qualifiers not stated in the text.

Representation rules:
- Use Python-style symbolic expressions.
- Materialize all tabular or graph data as explicit nested lists.
- Express all constraints as atomic expressions.

Output format (EXACT):
Return exactly one JSON object with the following structure:

{
  "problem_description": "...",
  "decision_variables": [
    { "name": "...", "type": "INTEGER/CONTINUOUS/BINARY", "description": "..." }
  ],
  "inputs": [
    { "name": "...", "value": "...", "units": "...", "description": "..." }
  ],
  "exogenous_variables": [],
  "exogenous_uncertainties": [],
  "state_variables": [],
  "transition_function": "",
  "objective_function": {
    "direction": "minimize/maximize",
    "expression": "...",
    "description": "..."
  },
  "constraints": [
    { "expression": "...", "description": "..." }
  ]
}

Inputs:
- Problem description: {problem_description}
\end{promptbox}
\vspace{1cm}
\FloatBarrier
\begin{promptbox}{MBR Candidate Re-ranking Prompt}
Act as an adjudicator selecting the single best Operations Research formulation from a small set of top candidates with subtle logical differences.

Inputs:
- Problem description: {problem_description}
- Candidate formulations (JSON): {candidates_json}

Decision protocol:
1. Identify functional differences (missing or extra constraints, inequality or objective direction changes, variable type mismatches).
2. Verify against the problem text:
   - Penalize unsupported or omitted constraints.
   - Disqualify candidates using prose instead of symbolic math.
   - Prefer balance equations over redundant static bounds when they implicitly enforce limits.
3. Select the mathematically correct, non-redundant formulation; prefer fewer, more general constraints when multiple candidates are valid.

Output:
Return ONLY a JSON object in the following format:

{
  "disagreement_analysis": "Brief summary of key conflicts.",
  "best_candidate_id": 1,
  "confidence": "high|medium|low",
  "reasoning": "Concise justification referencing correctness and avoidance of redundancy."
}

Constraints:
- best_candidate_id must match a provided candidate ID.
- No text outside the JSON object.
\end{promptbox}
\subsubsection{Mathematical Formulation}
\FloatBarrier
\begin{questionbox}{Problem Description: Humanitarian Food Distribution Scenario}
Imagine you are the director of a non-profit organization tasked with providing food supplies to six regions suffering from a famine. Each region has a certain amount of food already, but they require more to sustain their population through the hardship.

Here are the current quantities of food (in tons) available and the required quantities for each region:
\begin{itemize}
    \item Region 1 has 42 tons but needs 74 tons.
    \item Region 2 has 32 tons but needs 476 tons.
    \item Region 3 has 398 tons but only needs 2 tons.
    \item Region 4 has 224 tons but needs 235 tons.
    \item Region 5 has 210 tons but needs 221 tons.
    \item Region 6 has 209 tons but only needs 72 tons.
\end{itemize}

You have the ability to transfer food supplies from one region to another. However, the cost of transportation varies depending on which regions you are transferring food between. Below is a list detailing the cost of moving food from one region to any other:

\begin{itemize}
    \item To move food from/to Region 1: To Region 2 costs 16, to Region 3 costs 48, to Region 4 costs 42, to Region 5 costs 50, to Region 6 costs 8.
    \item To move food from/to Region 2: To Region 1 costs 27, to Region 3 costs 23, to Region 4 costs 37, to Region 5 costs 39, to Region 6 costs 29.
    \item To move food from/to Region 3: To Region 1 costs 49, to Region 2 costs 39, to Region 4 costs 33, to Region 5 costs 50, to Region 6 costs 6.
    \item To move food from/to Region 4: To Region 1 costs 23, to Region 2 costs 49, to Region 3 costs 46, to Region 5 costs 50, to Region 6 costs 6.
    \item To move food from/to Region 5: To Region 1 costs 45, to Region 2 costs 47, to Region 3 costs 48, to Region 4 costs 26, to Region 6 costs 39.
    \item To move food from/to Region 6: To Region 1 costs 33, to Region 2 costs 11, to Region 3 costs 9, to Region 4 costs 4, to Region 5 costs 12.
\end{itemize}

Your mission is to ensure every region receives the food it needs while keeping the transportation cost as low as possible. What would be the minimum cost to make sure all regions have enough food?
\end{questionbox}
\vspace{1cm}
\FloatBarrier
\begin{resultbox}{Mathematical Formulation}
{
  "decision_variables": [
    {
      "name": "x[i,j]",
      "type": "CONTINUOUS",
      "description": "tons of food shipped from region i to region j (i != j, i, j in {1,...,6})"
    }
  ],
  "inputs": [
    {
      "name": "current_food",
      "value": [42, 32, 398, 224, 210, 209],
      "units": "tons",
      "description": "current quantity of food available in each region i (i = 1,...,6)"
    },
    {
      "name": "required_food",
      "value": [74, 476, 2, 235, 221, 72],
      "units": "tons",
      "description": "required quantity of food for each region i (i = 1,...,6)"
    },
    {
      "name": "transportation_costs",
      "value": [
        [0, 16, 48, 42, 50, 8],
        [27, 0, 23, 37, 39, 29],
        [49, 39, 0, 33, 50, 6],
        [23, 49, 46, 0, 50, 6],
        [45, 47, 48, 26, 0, 39],
        [33, 11, 9, 4, 12, 0]
      ],
      "units": "cost units per ton",
      "description": "matrix where entry [i][j] is the cost of transporting one ton of food from region i+1 to region j+1"
    }
  ],
  "exogenous_variables": [],
  "exogenous_uncertainties": [],
  "state_variables": [],
  "transition_function": "",
  "objective_function": {
    "direction": "minimize",
    "expression": "sum(transportation_costs[i][j] * x[i,j] for i in regions for j in regions if i != j)",
    "description": "minimize total transportation cost of shipping food between regions"
  },
  "constraints": [
    {
      "expression": "current_food[i] + sum(x[j,i] for j in regions if j != i) - sum(x[i,j] for j in regions if j != i) >= required_food[i]  for all i in regions",
      "description": "each region must end up with at least its required amount of food"
    },
    {
      "expression": "x[i,j] >= 0  for all i, j in regions with i != j",
      "description": "non-negativity of shipped quantities"
    }
  ]
}
\end{resultbox}

\subsubsection{Component MBR Re-ranking}

In this stage, the system employs the two-step refinement process illustrated in Figure \ref{fig:mbr}. First, it generates $5$ candidate extractions and filters them using component-level consensus scores (Stage 1). Second, an LLM-based reranker analyzes the top 3 candidates to select the final output based on logical completeness (Stage 2).
\begin{resultbox}{Stage 1: Component-Level MBR Filtering}
{
  "total_candidates": 5,
  "filtering_logic": "Select Top-3 based on embedding consensus score",
  "scores": [
    { "id": 5, "score": 0.861, "status": "PASSED" },
    { "id": 2, "score": 0.860, "status": "PASSED" },
    { "id": 3, "score": 0.857, "status": "PASSED" },
    { "id": 4, "score": 0.785, "status": "FILTERED (Score too low)" },
    { "id": 1, "score": 0.776, "status": "FILTERED (Score too low)" }
  ]
}
\end{resultbox}
\vspace{1cm}
\FloatBarrier
\begin{resultbox}{Stage 2: LLM Re-ranking (Selection from Top 3)}
{
  "disagreement_analysis": "Main difference is the balance constraint. Candidates 2 and 3 impose strict equality: final stock must equal the exact requirement, expressed as (out-in) = current - required. Because total initial supply (1115) exceeds total need (1080) by 35 tons, this global equality system is infeasible (sum of left sides is 0, right side 35). Candidate 5 instead uses an inequality: current + in - out >= required, which matches the text ('make sure all regions have enough food') and allows leftover surplus. All candidates include non-negativity and the correct cost-minimisation objective; Candidate 2 also allows x[i][i] flows and omits the i!=j restriction. No other material disagreements.",
  "best_candidate_id": 5,
  "confidence": "high",
  "reasoning": "Candidate 5 is the only formulation that faithfully captures the requirement 'at least the needed amount' with a >= constraint, therefore remains feasible given the overall surplus. It omits self-ship variables, has the correct minimisation objective and enforces non-negativity. Candidates 2 and 3, by forcing exact equality, would make the model infeasible for the stated data, so they must be rejected."
}
\end{resultbox}
\vspace{1cm}
\begin{resultbox}{Constraint Comparison: Selected vs. Rejected}
{
  "Selected Formulation": {
    "constraints": [
      {
        "expression": "current_food[i] + sum(x[j,i] for j in regions if j != i) - sum(x[i,j] for j in regions if j != i) >= required_food[i] for all i in regions",
        "description": "each region must end up with at least its required amount of food (allows surplus)"
      }
    ]
  },
  "Rejected Formulation": {
    "constraints": [
      {
        "expression": "sum(x[i][j] for j in regions) - sum(x[j][i] for j in regions) == current_food[i] - required_food[i] for all i in regions",
        "description": "food balance for each region: net shipments out minus net shipments in equals its initial surplus or deficit (forces exact equality, causing infeasibility)"
      }
    ]
  }
}
\end{resultbox}
\subsection{Solver Recommender}
\FloatBarrier
\begin{promptbox}{Solver Recommender Prompt}
Inputs:
- A decision problem specified as KEY_INGREDIENTS: {key_ingredients}
- A candidate solver list SOLVERS: {solvers}

Goal:
Recommend a ranked list of solvers to solve the given problem.

System logic:
- Assume a working Gurobi commercial license is available, and mention this explicitly in the recommendation.
- Do not recommend open-source alternatives solely for licensing reasons, since Gurobi can be used without restrictions.
- Rank solvers primarily by:
  1) simplicity of use and integration,
  2) expected suitability for the problem size and complexity implied by KEY_INGREDIENTS.
- Provide recommendations only (no implementation details or code beyond what is required to communicate solver choice rationale).
\end{promptbox}
\vspace{1cm}
\FloatBarrier
\begin{resultbox}{Solver Ranking}

Recommended solvers for this LP-type “minimum-cost transportation” problem (ranked from most convenient to least):

----------------------------------------
1. gurobipy (Gurobi's Python API) [SELECTED]
----------------------------------------
Why:
- Native linear-programming engine; solves this six-node balanced transportation model in milliseconds.
- Extremely simple modeling interface (addVars, addConstrs, setObjective).
- You already have a commercial license and the environment variables GRB_WLSACCESSID, GRB_WLSSECRET, and GRB_LICENSEID are set, so you can start immediately.

How to install:
pip install gurobipy

How to activate license:
import os, gurobipy as gp
options = {
  'WLSACCESSID': os.environ.get('GRB_WLSACCESSID'),
  'WLSSECRET':   os.environ.get('GRB_WLSSECRET'),
  'LICENSEID':   int(os.environ.get('GRB_LICENSEID'))
}
env = gp.Env(params=options)

----------------------------------------
2. CVXPY (open-source, high-level convex modeling)
----------------------------------------
Why:
- Very concise (problem can be written in ~15 lines).
- Automatically picks the best open-source LP solver installed (default ECOS/CBC).

----------------------------------------
3. OR-Tools (Google)
----------------------------------------
Why:
- Contains a dedicated min-cost flow interface even higher-level than plain LP.
- Pure Python wheel, no external solver needed.

----------------------------------------
4. Pyomo / 5. SciPy / 6. pyscipopt
----------------------------------------
(Listed as alternatives but ranked lower due to verbosity or unnecessary overhead for this specific small LP.)

========================================
Summary
========================================
- Use gurobipy first—fastest path to an optimal answer, license already configured.
- CVXPY and OR-Tools are the simplest open-source fall-backs.

\end{resultbox}
\subsection{Simulator}
\begin{figure}[h]
\vskip 0.2in
\centering
\begin{tikzpicture}[
    font=\sffamily,
    folder/.style={
        draw=gray!45,
        rectangle,
        rounded corners=2pt,
        minimum width=2.6cm,
        minimum height=0.45cm,
        fill=blue!30,
        font=\scriptsize\ttfamily,
        drop shadow={opacity=0.18, shadow xshift=1pt, shadow yshift=-1pt}
    },
    file/.style={
        draw=gray!55,
        line width=0.6pt,
        rectangle,
        minimum width=2.4cm,
        minimum height=0.35cm,
        fill=white,
        font=\tiny\ttfamily,
        inner sep=2pt
    },
    arrow/.style={->, >=stealth, thick},
    tree line/.style={draw=gray!45, thick},
    test line/.style={draw=gray!60, densely dotted, thick},
    node distance=0.35cm,
]


\node[folder, fill=blue!30, minimum width=10.6cm]
(workspace) at (0, 3.25) {/workspace/nemo/};

\node[folder, fill=green!35] (examples) at (-4.1, 2.05) {examples/};
\node[file] (ex1) at (-4.1, 1.55) {example\_1.py};
\node[file] (ex2) at (-4.1, 1.15) {example\_2.py};
\node[file] (ex3) at (-4.1, 0.75) {example\_3.py};

\node[font=\scriptsize\bfseries, text=green!55!black, align=center]
(lbl_ref) at (-4.1, 0.20) {Few-shot\\[-1pt]Reference};

\node[folder, fill=orange!35] (simulator) at (0, 2.05) {simulator\_code/};
\node[file] (sim1) at (0, 1.55) {models.py};
\node[file] (sim2) at (0, 1.15) {constraints.py};
\node[file] (sim3) at (0, 0.75) {objective.py};

\node[folder, fill=orange!28] (sim_tests) at (0, 0.15) {simulator\_tests/};
\node[file] (simt1) at (0, -0.25) {test\_simulator.py};

\node[folder, fill=purple!35] (optimizer) at (4.1, 2.05) {optimizer\_code/};
\node[file] (opt1) at (4.1, 1.65) {variant\_1/};
\node[file] (opt2) at (4.1, 1.25) {variant\_2/};
\node[file] (opt3) at (4.1, 0.85) {variant\_3/};
\node[file] (opt4) at (4.1, 0.45) {ensemble.py};

\node[folder, fill=purple!28] (opt_tests) at (4.1, -0.15) {optimizer\_tests/};
\node[file] (optt1) at (4.1, -0.55) {test\_optimizer.py};

\draw[tree line] (workspace.south) -- +(0,-0.35) -| (examples.north);
\draw[tree line] (workspace.south) -- +(0,-0.35) -| (optimizer.north);
\draw[tree line] (workspace.south) -- (simulator.north);

\draw[tree line] (examples.south) -- (ex1.north);
\draw[tree line] (simulator.south) -- (sim1.north);
\draw[tree line] (optimizer.south) -- (opt1.north);

\draw[test line] (sim3.south) -- (sim_tests.north);
\draw[test line] (opt4.south) -- (opt_tests.north);

\draw[tree line] (sim_tests.south) -- (simt1.north);
\draw[tree line] (opt_tests.south) -- (optt1.north);

\node[font=\scriptsize\bfseries, text=orange!70!black, align=center]
(lbl_chk) at (0, -1.1) {Feasibility\\[-1pt]Checker};

\node[font=\scriptsize\bfseries, text=purple!70!black, align=center]
(lbl_sol) at (4.1, -1.1) {Solution\\[-1pt]Generator};

\coordinate (arrow_top_y) at ($(lbl_chk.south) + (0, 0.3)$);
\coordinate (arrow_bot_y) at ($(lbl_chk.south) + (0, -0.15)$);

\coordinate (chk_top) at (lbl_chk.east |- arrow_top_y);
\coordinate (sol_top) at (lbl_sol.west |- arrow_top_y);
\coordinate (sol_bot) at (lbl_sol.west |- arrow_bot_y);
\coordinate (chk_bot) at (lbl_chk.east |- arrow_bot_y);

\draw[arrow, orange!80!black, shorten <=2pt, shorten >=2pt]
    (chk_top) -- node[above, font=\tiny, text=orange!80!black]
    {validates} (sol_top);

\draw[arrow, blue!75!black, shorten <=2pt, shorten >=2pt]
    (sol_bot) -- node[below, font=\tiny, text=blue!75!black]
    {proposes $x^*$} (chk_bot);

\end{tikzpicture}
\caption{Coding agent workspace structure. Retrieved examples are materialized as executable Python files in \texttt{examples/}. The agent independently generates a simulator and an optimizer; the optimizer contains three independent implementations of the optimizer. Associated test suites in \texttt{simulator\_tests/} and \texttt{optimizer\_tests/} provide regression, feasibility, and solver-consistency checks.}
\label{fig:workspace}
\end{figure}
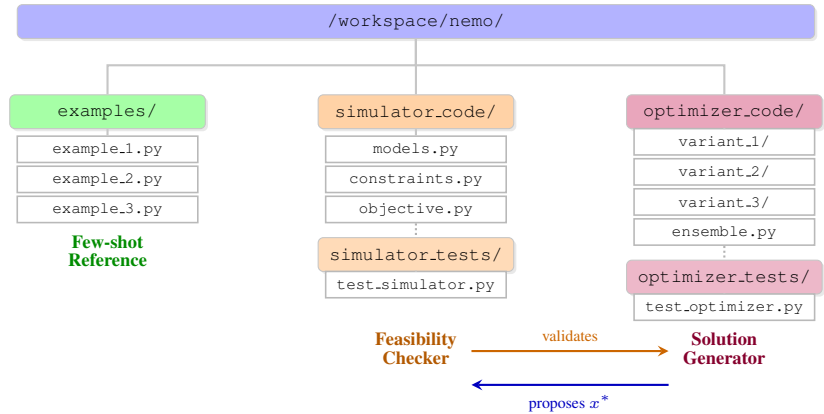

As depicted in the \texttt{simulator\_code/} branch of Figure \ref{fig:workspace}, the coding agent first materializes a simulation environment. Generated via a single prompt provided below, this module acts as the independent validator for any proposed solutions.

The core logic resides in \texttt{constraints.py}, which enforces the physical rules of the system (e.g., flow conservation and non-negativity). A snippet of this generated verification logic is provided below.
\FloatBarrier
\begin{promptbox}{Simulator Creation Prompt}
Define a simulator (not an optimizer) for a decision process specified by a structured KEY_INGREDIENTS JSON, including the full natural-language problem_description.

System logic:
- The simulator represents the mechanics of the decision process exactly as specified in KEY_INGREDIENTS.
- It provides explicit representations for:
  - decision variables
  - system state
  - exogenous inputs
  - uncertain variables
- Given a concrete set of inputs, the simulator:
  1) checks all constraints and reports violations,
  2) evaluates the objective and related metrics only when constraints are satisfied,
  3) applies the transition function to update system state when applicable.
- The simulator evaluates provided decisions only and does not perform optimization, search, or inference of optimal actions.
- The simulator must not assume optimality, solver behavior, or hidden objectives beyond what is stated.

Architectural separation:
- The simulator is a standalone system component and is logically isolated from any optimizer or solver logic.
- It serves as a ground-truth evaluator for decisions, not a decision generator.

Directory structure (architectural contract):
/workspace/nemo/
|-- simulator_code/        (simulator logic)
|-- simulator_tests/       (validation and diagnostics)

Inputs:
- Decision process specification: {key_ingredients}
\end{promptbox}

\vspace{1cm}
\FloatBarrier
\begin{resultbox}{Simulator Implementation (\texttt{constraints.py})}
def check_all_constraints(decision, inputs):
    violations = []
    
    # 1. Check Non-negativity
    if (v := check_non_negativity_constraint(decision)):
        violations.append(v)
    
    # 2. Check Food Requirements (Flow Balance)
    violations.extend(check_food_requirement_constraint(decision, inputs))
    
    return violations

def check_food_requirement_constraint(decision, inputs):
    """
    Check that each region ends up with at least its required amount of food.
    Logic: Final = Initial + Inflow - Outflow >= Required
    """
    violations = []
    
    # Start with current food (copy to avoid mutating input)
    net_food = inputs.current_food.copy()
    
    # Apply all shipments
    for (from_region, to_region), amount in decision.shipments.items():
        net_food[from_region - 1] -= amount # Outflow
        net_food[to_region - 1] += amount   # Inflow
    
    # Check if each region has enough food
    for region in range(1, 7):
        idx = region - 1
        if net_food[idx] < inputs.required_food[idx]:
            violations.append(ConstraintViolation(
                constraint_name="food_requirement",
                description=f"Region {region} insufficient food",
                details=f"Has {net_food[idx]:.2f} tons, needs {inputs.required_food[idx]}"
            ))
    
    return violations
\end{resultbox}
\subsection{Optimizer}
\subsubsection{Prompts}
\FloatBarrier
\begin{promptbox}{Optimizer Creator Prompt}
Solve a single optimization problem defined by KEY_INGREDIENTS and produce a valid solver-derived result.

Core logic:
- Interpret the problem strictly and literally from KEY_INGREDIENTS, including the full problem_description.
- Build an optimization model that exactly matches the specified decision variables, constraints, and objective.
- Do not introduce additional constraints, objectives, assumptions, or problem modifications.
- Solve the problem using one of the provided LIBRARY/SOLVER options.
- If multiple solvers are provided, attempt them in ranked order until a valid result is obtained.
- Results must be produced by an actual solver run; fabricated or assumed solutions are forbidden.

System invariants:
- The simulator is treated as immutable and must not be modified or relied upon during optimization.
- The optimizer solves only the given problem instance (no toy data or alternate inputs).
- Solver status must be reported truthfully (optimal, infeasible, unbounded, time_limit, or error).

Directory structure (logical contract):
The optimizer exists as a separate module alongside the simulator, with optimizer logic isolated from simulator logic:

/workspace/nemo/
|-- simulator_code/        (immutable; not used by optimizer)
|-- simulator_tests/       (immutable)
|-- optimizer_code/        (optimizer implementation)
|-- optimizer_tests/       (optimizer outputs and validation artifacts)

Output format (EXACT):
Produce a JSON object with the following structure:

{
  "optimal_variables": {"var_name": value, ...},
  "optimal_objective_value": float,
  "status": "optimal|infeasible|unbounded|time_limit|error",
  "solver_info": {
    "solver_name": "string",
    "solve_time": float,
    "iterations": int,
    "gap": float
  }
}

Inputs:
- Optimization problem: {key_ingredients}
- Solver recommendations: {library_recommendation}
\end{promptbox}

\vspace{1cm}
\FloatBarrier
\begin{promptbox}{Optimizer Self-Consistency Prompt}
Solve a single optimization problem using an ensemble of {num_variants} independent optimizer variants and produce a consensus solution.

ENSEMBLE LOGIC:
- Create {num_variants} independent optimizer variants.
- All variants solve the SAME problem defined by KEY_INGREDIENTS.
- Variants are independent implementations to reduce modeling and implementation errors.
- Each variant produces its own optimization result.
- Final solution is obtained via majority voting across variants.

CONSENSUS RULES:

Status consensus:
- Count solver statuses across variants.
- Select the most frequent status.
- Break ties using the priority order:
  optimal > time_limit > infeasible > unbounded > error

Objective value consensus:
- Group objective values that agree within tolerance:
  * relative tolerance: 1e-6
  * absolute tolerance: 1e-9
- Select the objective value with the largest agreement group.
- If tied, select the median of the tied group.

Variable consensus:
- Identify variants that produced the majority objective value.
- Select decision variables from one agreeing variant.
- Do NOT average integer or binary variables.

Tracking:
- Track successful vs failed variants.
- Track solver agreement and objective agreement.
- Record variants that produced outlier or failed results.

Variant output format (EXACT):
Each variant outputs:

{
  "optimal_variables": {"var_name": value, ...},
  "optimal_objective_value": float,
  "status": "optimal|infeasible|unbounded|time_limit|error",
  "solver_info": {
    "solver_name": "string",
    "solve_time": float,
    "iterations": int,
    "gap": float
  }
}

FINAL ENSEMBLE OUTPUT FORMAT (EXACT):
The ensemble produces a consensus result:

{
  "optimal_variables": {"var_name": consensus_value, ...},
  "optimal_objective_value": majority_objective_value,
  "status": "majority_vote_status",
  "solver_info": {
    "ensemble_size": {num_variants},
    "solvers_used": ["solver names"],
    "consensus_solvers": ["solvers agreeing"]
  },
  "consensus_info": {
    "num_variants": int,
    "num_successful": int,
    "num_failed": int,
    "status_distribution": {},
    "solver_agreement": int,
    "objective_agreement": int,
    "objective_agreement_ratio": float,
    "num_unique_objectives": int,
    "failed_variants": []
  },
  "variant_results": [
    {
      "variant_name": "variant_1",
      "solver": "solver_name",
      "status": "optimal",
      "objective_value": float,
      "solve_time": float
    }
  ]
}
\end{promptbox}

\vspace{1cm}
\FloatBarrier
\begin{promptbox}{Asymmetric Validation Prompt}
Validate optimizer results against the simulator. Modify optimizer code and re-run only if validation fails.

Rules:
- No mocked solutions or manual edits to optimization_results.json.
- If re-running is needed, execute the real optimizer script.
- Validation code must live under optimizer_tests/.

Process:
1. Load optimizer_tests/optimization_results.json.
2. Validate results using the simulator (read-only).
3. If valid, exit. If invalid, fix code, re-run optimizer, and re-validate (max 3 iterations).

Checks:
- Verify optimal_variables and dimensions.
- Validate constraints and bounds via the simulator.
- Recompute objective and compare (tolerance 1e-4).
- Flag trivial all-zero optimal solutions.
- Identify responsible optimizer files and required fixes.

Output format (EXACT):
Write optimizer_tests/validation_results.json as:

{
  "passed": bool,
  "num_validation_iterations": int,
  "problem_analysis": {
    "problem_feasible": bool,
    "has_trivial_solutions": bool,
    "infeasibility_analysis": {}
  },
  "input_verification": {},
  "constraint_violations": [],
  "bound_violations": [],
  "objective_verification": {
    "optimizer_value": float,
    "simulator_value": float,
    "difference": float,
    "match": bool
  },
  "validation_history": [
    {
      "iteration": int,
      "passed": bool,
      "issues_found": [],
      "fixes_applied": []
    }
  ]
}
\end{promptbox}

\subsubsection{Generated Optimizer Code}

Moving to the \texttt{optimizer\_code/} branch of Figure \ref{fig:workspace}, the coding agent generates three independent solver implementations (Variants 1–3) to enable self-consistency checking. Below we present the code for Variant 1, which utilizes the Gurobi solver as recommended.

Additionally, we capture the full agent interaction history using the \emph{OpenHands} Trajectory API. This allows us to trace the agent's debugging steps—specifically how it corrects syntax errors or formulation bugs—providing a granular audit trail for the system's reasoning process.
\FloatBarrier
\begin{resultbox}{Optimizer Implementation (Variant 1 - Gurobi)}
# --- optimizer.py (Main Driver) ---
class FoodDistributionOptimizer:
    def build_model(self) -> None:
        # 1. Create Model
        self.model = gp.Model("FoodDistribution", env=self.env)
        
        # 2. Create Variables (x[i,j])
        self.x_vars = {}
        for i in range(1, self.num_regions + 1):
            for j in range(1, self.num_regions + 1):
                if i != j:  # No shipments to self
                    self.x_vars[(i, j)] = self.model.addVar(
                        lb=0.0,  # Non-negativity constraint
                        name=f"x_{i}_{j}"
                    )
        
        # 3. Add Constraints & Objective
        add_all_constraints(self.model, self.x_vars, self.inputs)
        set_objective_function(self.model, self.x_vars, self.inputs)
        
        self.model.update()

    def solve(self, time_limit_seconds=None):
        if self.model is None:
            self.build_model()
        if time_limit_seconds:
            self.model.setParam('TimeLimit', time_limit_seconds)
            
        self.model.optimize()
        return self._process_results(self.model.Runtime)

# --- constraints.py (Mathematical Logic) ---
def add_food_requirement_constraints(model, x_vars, inputs):
    """
    Enforces: current_food[i] + Inflow - Outflow >= required_food[i]
    """
    num_regions = len(inputs.current_food)
    
    for i in range(num_regions):
        region = i + 1  # 1-based indexing
        
        # Calculate incoming: sum(x[j,i] for j != i)
        incoming = gp.quicksum(
            x_vars.get((j+1, region), 0) 
            for j in range(num_regions) if j+1 != region
        )
        
        # Calculate outgoing: sum(x[i,j] for j != i)
        outgoing = gp.quicksum(
            x_vars.get((region, j+1), 0) 
            for j in range(num_regions) if j+1 != region
        )
        
        # Add constraint (ALLOWS surplus food)
        model.addConstr(
            inputs.current_food[i] + incoming - outgoing >= inputs.required_food[i],
            name=f"food_requirement_region_{region}"
        )
\end{resultbox}
\subsubsection{Optimizer Results \& Validation}

Following code generation, the system executes the \texttt{ensemble.py} script (see Figure \ref{fig:workspace}). This orchestrator triggers all three generated solver variants in parallel and aggregates their results to verify mathematical consensus.

The output below shows the exact JSON structure returned by this ensemble execution, confirming that the agent correctly formatted the response and achieved unanimous agreement on the objective value.
\FloatBarrier
\begin{resultbox}{Optimizer Results}
{
  "optimal_variables": {
    "0,1": 0.0, "0,2": 0.0, "0,3": 0.0, "0,4": 0.0, "0,5": 0.0,
    "2,5": 361.0,  // ... (Non-zero active route)
    "3,0": 32.0,   // ... (Non-zero active route)
    "5,1": 444.0,  // ... (Non-zero active route)
    "5,3": 43.0,   // ... (Non-zero active route)
    "5,4": 11.0    // ... (Non-zero active route)
    // ... (Remaining zero-flow variables omitted for brevity)
  },
  "optimal_objective_value": 8090.0,
  "status": "optimal",
  "solver_info": {
    "ensemble_size": 3,
    "solvers_used": [ "Gurobi", "CVXPY (ECOS)", "OR-Tools (GLOP)" ],
    "consensus_solvers": [ "Gurobi", "CVXPY (ECOS)", "OR-Tools (GLOP)" ]
  },
  "consensus_info": {
    "num_variants": 3,
    "num_successful": 3,
    "num_failed": 0,
    "status_distribution": { "optimal": 3 },
    "solver_agreement": 3,
    "objective_agreement": 3,
    "objective_agreement_ratio": 1.0,
    "num_unique_objectives": 1,
    "failed_variants": []
  },
  "variant_results": [
    {
      "variant_name": "variant_1",
      "solver": "Gurobi",
      "status": "optimal",
      "objective_value": 8090.0,
      "solve_time": 0.0019
    },
    {
      "variant_name": "variant_2",
      "solver": "CVXPY (ECOS)",
      "status": "optimal",
      "objective_value": 8090.00000015,
      "solve_time": 0.0359
    },
    {
      "variant_name": "variant_3",
      "solver": "OR-Tools (GLOP)",
      "status": "optimal",
      "objective_value": 8090.0,
      "solve_time": 0.0016
    }
  ]
}
\end{resultbox}
\vspace{1cm}

Finally, the system validates the proposed solution against the simulator. As shown in the validation log below, the simulator independently verifies that the solution satisfies all constraints (returns an empty violation list) and that the re-calculated objective value matches the optimizer's report exactly.
\FloatBarrier
\begin{resultbox}{Validation Results}
{
  "passed": true,
  "validation_history": [
    {
      "iteration": 0,
      "passed": true,
      "issues_found": [],
      "fixes_applied": []
    }
  ],
  "problem_analysis": {
    "problem_feasible": true,
    "has_trivial_solutions": false
  },
  "constraint_violations": [],  // Empty list confirms physical feasibility
  "objective_verification": {
    "passed": true,
    "optimizer_value": 8090.0,
    "simulator_value": 8090.0,
    "difference": 0.0,          // Perfect alignment with the simulator
    "issues": []
  }
}
\end{resultbox}
\subsubsection{Retrieved Few-Shot Examples}
As a preliminary step before code generation, the system retrieves relevant solved instances from the vectorstore based on semantic similarity. These samples are uploaded directly into the \emph{OpenHands} workspace (specifically the \texttt{examples/} directory shown in Figure \ref{fig:workspace}), providing the agent with concrete reference implementations. Below is one such retrieved artifact.
\FloatBarrier
\begin{resultbox}{Retrieved Code Artifact}
"""
Training Example 1 | Similarity: 0.716 | Type: transportation
Question: Managing Food Distribution Across Six Cities
... (Problem Description: surplus/deficits and shipment costs) ...
"""

import gurobipy as gp
from gurobipy import GRB

# Data
cities = [0, 1, 2, 3, 4, 5]
supply = {0: 17, 1: 5, 2: 3, 3: -7, 4: -21, 5: 3}
cost = {
    (0, 1): 3, (0, 2): 6, (0, 3): 4, (0, 4): 8, (0, 5): 5,
    # ... (truncated for brevity)
}
capacity = {
    (0, 1): 32, (0, 2): 16, (0, 3): 38, (0, 4): 14, (0, 5): 34,
    # ... (truncated for brevity)
}

# Model
model = gp.Model("FoodDistribution")
x = model.addVars(cities, cities, name="x")

# Objective
model.setObjective(gp.quicksum(cost[i,j]*x[i,j] for i,j in cost), GRB.MINIMIZE)

# Constraints
for i in cities:
    # Supply/Demand Balance
    model.addConstr(
        gp.quicksum(x[i,j] for j in cities) - 
        gp.quicksum(x[j,i] for j in cities) == supply[i]
    )

for i,j in capacity:
    model.addConstr(x[i,j] <= capacity[i,j])

model.optimize()
\end{resultbox}
\end{document}